\RequirePackage{fix-cm}
\documentclass[twocolumn]{svjour3}          
\smartqed  
\usepackage{graphicx}

\usepackage{amsmath,amssymb} 
\DeclareMathOperator*{\argmax}{argmax} 

\usepackage{color}

\usepackage[acronym,toc]{glossaries} 
\usepackage[toc,page]{appendix}

\usepackage{tikz}
\usepackage[utf8]{inputenc}
\usepackage{pgfplots}
\usepackage{subfig}

\usepackage{natbib}

\usepackage{booktabs}
\usepackage{adjustbox} 
\usepackage{url}
\usepackage[
linkcolor						= blue,
citecolor						= green!90!black,
colorlinks   = true
]{hyperref}

\definecolor{darkgreen}{rgb}{0.498,0.749,0.498}
\definecolor{darkyellow}{rgb}{1.0,0.89,0.498}
\definecolor{darkred}{rgb}{1.0,0.498,0.498}
\usepackage{pifont}
\newcommand{\OK}{\ding{51}}

\newglossaryentry{VAE}{name=VAE,description={Variational Autoencoder},first={\glsentrydesc{VAE} (\glsentrytext{VAE})}}
\newglossaryentry{AE}{name={AE},description={Autoencoder},first={\glsentrydesc{AE} (\glsentrytext{AE})},plural=AEs,descriptionplural={Autoencoders},firstplural={\glsentrydescplural{AE} (\glsentryplural{AE})}}
\newglossaryentry{AAE}{name={AAE},description={Augmented Autoencoder},first={\glsentrydesc{AAE} (\glsentrytext{AAE})},plural=AAEs,descriptionplural={Augmented Autoencoders},firstplural={\glsentrydescplural{AAE} (\glsentryplural{AAE})}}
\newglossaryentry{CVAE}{name=CVAE,description={Conditional Variational Autoencoder},first={\glsentrydesc{CVAE} (\glsentrytext{CVAE})}}
\newglossaryentry{SSD}{name={SSD},description={Single Shot Multibox Detector},first={\glsentrydesc{SSD} (\glsentrytext{SSD})}}
\newglossaryentry{CNN}{name=CNN,description={Convolutional Neural Network},first={\glsentrydesc{CNN} (\glsentrytext{CNN})}, plural=CNNs,descriptionplural={Convolutional Neural Networks},firstplural={\glsentrydescplural{CNN} (\glsentryplural{CNN})}}
\newglossaryentry{SIFT}{name=SIFT,description={Scaled Invariant Feature Transform},first={Scaled Invariant Feature Transform (SIFT)}}
\newglossaryentry{SURF}{name=SURF,description={Speeded Up Robust Features},first={Speeded Up Robust Features (SURF)}}
\newglossaryentry{PnP}{name=PnP,description={Perspective-n-Point}, first={Perspective-n-Point (PnP)}}
\newglossaryentry{RF}{name=RF,description={Random Forrest},first={Random Forest (RF)},plural=RFs,descriptionplural={Random Forests},
	firstplural={\glsentrydescplural{RF} (\glsentryplural{RF})}}
\newglossaryentry{RANSAC}{name=RANSAC,description={Random Sample Consensus},first={RANSAC},}
\newglossaryentry{PCA}{name=PCA,description={Principal Component Analysis},first={Principal Component Analysis (PCA)}}
\newglossaryentry{LIDAR}{name={LIDAR},description={Light Detection And Ranging},first={\glsentrydesc{LIDAR} (\glsentrytext{LIDAR})}}
\newglossaryentry{kNN}{name={kNN},description={k-Nearest-Neighbor},first={\glsentrydesc{kNN} (\glsentrytext{kNN})}} 
\newglossaryentry{MLP}{name={MLP},description={Multilayer Perceptron},first={\glsentrydesc{MLP} (\glsentrytext{MLP})}, plural=MLPs,descriptionplural={Multilayer Perceptrons},firstplural={\glsentrydescplural{MLP} (\glsentryplural{MLP})}}
\newglossaryentry{EM}{name={EM},description={Expectation Maximization},first={\glsentrydesc{EM} (\glsentrytext{EM})}}
\newglossaryentry{6DOF}{name={6DOF},description={six degrees of freedom},first={\glsentrydesc{6DOF} (\glsentrytext{6DOF})}}
\newglossaryentry{ICP}{name={ICP},description={Iterative Closest Point},first={\glsentrydesc{ICP} (\glsentrytext{ICP})}}
\newglossaryentry{KL}{name={KL},description={Kullback-Leibler},first={\glsentrydesc{KL} (\glsentrytext{KL})}}
\newglossaryentry{VSD}{name={$err_{vsd}$},description={Visible Surface Discrepancy},first={\glsentrydesc{VSD} (\glsentrytext{VSD})}}
\newglossaryentry{mAP}{name={mAP},description={mean Average Precision},first={\glsentrydesc{mAP} (\glsentrytext{mAP})}}
\newglossaryentry{DA}{name={DA},description={Domain Adaptation},first={\glsentrydesc{DA} (\glsentrytext{DA})}}
\newglossaryentry{DR}{name={DR},description={Domain Randomization},first={\glsentrydesc{DR} (\glsentrytext{DR})}}
\newglossaryentry{GAN}{name={GAN},description={Generative Adversarial Network},first={\glsentrydesc{GAN}(\glsentrytext{GAN})},plural=GANs,descriptionplural={Generative Adversarial Networks},firstplural={\glsentrydescplural{GAN} (\glsentryplural{GAN})}}
\newglossaryentry{PPF}{name={PPF},description={Point Pair Features},first={\glsentrydesc{PPF} (\glsentrytext{PPF})}}

\begin{document}

\title{Augmented Autoencoders: Implicit 3D Orientation Learning for 6D Object Detection 
}



\author{Martin Sundermeyer$^1$ \and Zoltan-Csaba Marton$^1$ \and Maximilian Durner$^1$  \and Rudolph Triebel$^{1,2}$}



\institute{$^1$German Aerospace Center (DLR), 82234 Wessling, Germany \\
	$^2$Technical University of Munich, 80333 Munich, Germany \\
	\email{\{martin.sundermeyer, zoltan.marton, \\maximilian.durner, rudolph.triebel\}@dlr.de}
	}

\date{Received: date / Accepted: date}
\maketitle

\begin{abstract}
	We propose a real-time RGB-based pipeline for object detection and 6D pose estimation. Our novel 3D orientation estimation is based on a variant of the Denoising Autoencoder that is trained on simulated views of a 3D model using Domain Randomization. 
	
	This so-called Augmented Autoencoder has several advantages over existing methods: It does not require real, pose-annotated training data, generalizes to various test sensors and inherently handles object and view symmetries. Instead of learning an explicit mapping from input images to object poses, it provides an implicit representation of object orientations defined by samples in a latent space. Our pipeline achieves state-of-the-art performance on the T-LESS dataset both in the RGB and RGB-D domain. We also evaluate on the LineMOD dataset where we can compete with other synthetically trained approaches.
	
	We further increase performance by correcting 3D orientation estimates to account for perspective errors when the object deviates from the image center and show extended results. Our code is available here 	
	\footnotemark \footnotetext{\url{https://github.com/DLR-RM/AugmentedAutoencoder}}
	
\end{abstract}

\section{Introduction}
One of the most important components of modern computer vision systems
for applications such as mobile robotic manipulation and augmented
reality is a reliable and fast 6D object detection module. 
Although, there are very encouraging recent results from \cite{xiang2017posecnn,kehl2017ssd,hodan2017tless,wohlhart2015learning,vidal20186d,hinterstoisser2016going,tremblay2018deep}, a general, easily applicable, robust and fast solution is not available, yet. The
reasons for this are manifold. First and foremost, current solutions
are often not robust enough against typical challenges such as object
occlusions, different kinds of background clutter, and dynamic changes
of the environment. Second, existing methods often require certain
object properties such as enough textural
surface structure or an asymmetric shape to avoid
confusions. And finally, current systems are not efficient in terms
of run-time and in the amount and type of annotated training data they require.

\begin{figure*}[t]
	\centering
	\captionsetup{width=\textwidth}
	\includegraphics[width=0.805\textwidth]{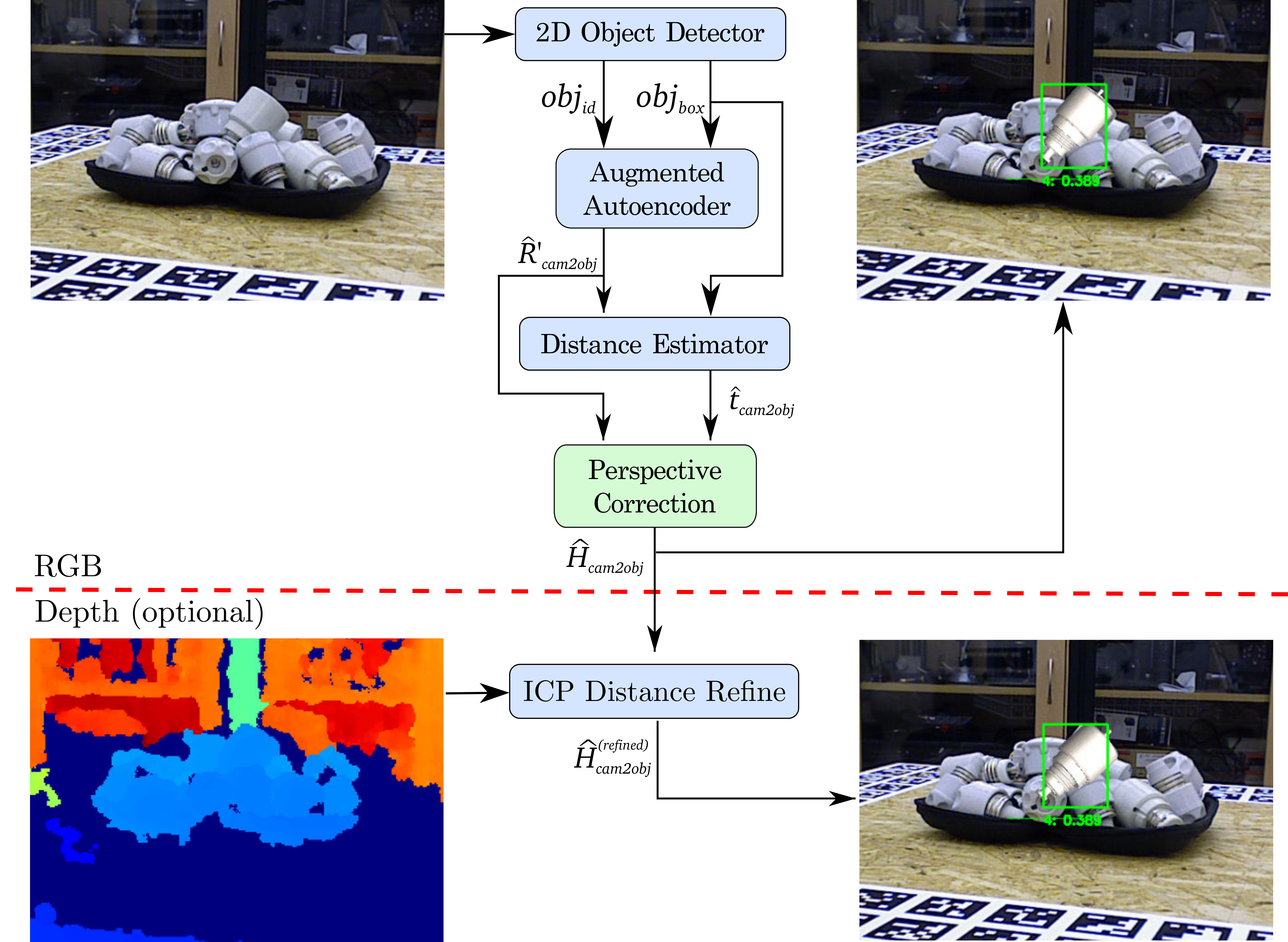}
	\caption{Our full 6D Object Detection pipeline: after detecting an object (2D Object Detector), the object is quadratically cropped and forwarded into the proposed Augmented Autoencoder. In the next step, the bounding box scale ratio at the estimated 3D orientation $\hat{R'}_{obj2cam}$ is used to compute the 3D translation $\hat{t}_{obj2cam}$. The resulting euclidean transformation $\hat{H'}_{obj2cam}  \in \mathcal{R}^{4x4}$ already shows promising results as presented in \cite{sundermeyer2018implicit}, however it still lacks of accuracy given a translation in the image plane towards the borders. Therefore, the pipeline is extended by the Perspective Correction block which addresses this problem and results in more accurate 6D pose estimates $\hat{H}_{obj2cam}$ for objects which are not located in the image center. Additionally, given depth data, the result can be further refined ($\hat{H}_{obj2cam}^{(refined)}$) by applying an Iterative Closest Point post-processing (bottom).}	\label{fig:pipeline}
\end{figure*}

Therefore, we propose a novel approach that directly
addresses these issues. Concretely, our method operates on single
RGB images, which significantly increases the usability as no depth
information is required. We note though that depth maps may be
incorporated optionally to refine the estimation. As a first
step, we build upon state-of-the-art 2D Object Detectors of (\cite{liu2016ssd,lin2018focal}) which provide object bounding boxes and identifiers. On the resulting scene crops, we
employ our novel 3D orientation estimation algorithm, which is based on a
previously trained deep network architecture. While deep networks are
also used in existing approaches, our approach differs in that we do
not explicitly learn from 3D pose annotations during training. Instead, we
\emph{implicitly} learn representations from rendered 3D model views. This is accomplished by training a generalized version of the Denoising Autoencoder from \cite{vincent2010stacked}, that we call \textit{'\gls{AAE}'}, using a novel Domain Randomization strategy. 
Our approach has several
advantages: First, since the training is independent from concrete representations of object orientations within $SO(3)$ (e.g. 
quaternions), we can handle ambiguous poses caused by symmetric views because we avoid one-to-many mappings from images to orientations. Second, we learn representations that specifically encode 3D orientations while achieving
robustness against occlusion, cluttered backgrounds and generalizing to different environments and test sensors. Finally, the \gls{AAE} does not require any real pose-annotated training data. Instead, it is trained to encode 3D model views in a self-supervised way, overcoming the need of a large pose-annotated dataset. A schematic overview of the approach based on \cite{sundermeyer2018implicit} is shown in Fig \ref{fig:pipeline}.

\section{Related Work\label{sec:relwork}}

Depth-based methods (e.g. using \gls{PPF} from \cite{vidal20186d,hinterstoisser2016going}) have shown robust pose estimation performance on multiple datasets, winning the SIXD challenge \citep{sixd,hodan2018bop}. However, they usually rely on the computationally expensive evaluation of many pose hypotheses and do not take into account any high level features. Furthermore, existing depth sensors are often more sensitive to sunlight or specular object surfaces than RGB cameras.

\glspl{CNN} have revolutionized 2D object detection from RGB images \citep{ren2015faster,liu2016ssd,lin2018focal}. But, in comparison to 2D bounding box annotation, the effort of labeling real images with full 6D object poses is magnitudes higher, requires expert knowledge and a complex setup \citep{hodan2017tless}. 

Nevertheless, the majority of learning-based pose estimation methods, namely \cite{tekin2017real,wohlhart2015learning,brachmann2016uncertainty,rad2017bb8, xiang2017posecnn}, use real labeled images that you only obtain within pose-annotated datasets. 

In consequence, \cite{kehl2017ssd, wohlhart2015learning, tremblay2018deep, zakharov2019dpod} have proposed to train on synthetic images rendered from a 3D model, yielding a great data source with pose labels free of charge. However, naive training on synthetic data does not typically generalize to real test images. Therefore, a main challenge is to bridge the domain gap that separates simulated views from real camera recordings. 

\subsection{Simulation to Reality Transfer}

There exist three major strategies to generalize from synthetic to real data: 

\subsubsection{Photo-Realistic Rendering} The works of \cite{movshovitz2016useful,su2015render,mitash2017self,richter2016playing} have shown that photo-realistic renderings of object views
and backgrounds can in some cases benefit the generalization performance for tasks like object detection and viewpoint estimation. It is especially suitable in simple environments and performs well if jointly trained with a relatively small amount of real annotated images.  However, photo-realistic modeling is often imperfect and requires much effort. Recently, \cite{Hodan2019PhotorealisticIS} have shown promising results for 2D Object Detection trained on physically-based renderings.

\subsubsection{Domain Adaptation} \gls{DA} \citep{csurka2017domain} refers to leveraging training data from a source domain to a target domain of which a small portion of labeled data (supervised \gls{DA}) or unlabeled data (unsupervised \gls{DA}) is available. \glspl{GAN} have been deployed for unsupervised \gls{DA} by generating realistic from synthetic images to train classifiers \citep{shrivastava2017learning}, 3D pose estimators \citep{bousmalis2017unsupervised} and grasping algorithms \citep{bousmalis2017using}. While constituting a promising approach, \glspl{GAN} often yield fragile training results. 
Supervised \gls{DA} can lower the need for real annotated data, but does not abstain from it. 

\subsubsection{Domain Randomization} \gls{DR} builds upon the hypothesis that by training a model on rendered views in a variety of semi-realistic settings (augmented with random lighting conditions, backgrounds, saturation, etc.), it will also generalize to real images. \cite{tobin2017domain} demonstrated the potential of the \gls{DR} paradigm for 3D shape detection using \glspl{CNN}. \cite{hinterstoisser2017pre} showed that by training only the head network of FasterRCNN of \cite{ren2015faster} with randomized synthetic views of a textured 3D model, it also generalizes well to real images. It must be noted, that their rendering is almost photo-realistic as the textured 3D models have very high quality. \cite{kehl2017ssd} pioneered an end-to-end \gls{CNN}, called 'SSD6D', for 6D object detection that uses a moderate \gls{DR} strategy to utilize synthetic training data. The authors render views of textured 3D object reconstructions at random poses on top of MS COCO background images \citep{lin2014microsoft} while varying brightness and contrast. This lets the network generalize to real images and enables 6D detection at 10Hz. Like us, for accurate distance estimation they rely on \gls{ICP} post-processing using depth data. In contrast, we do not treat 3D orientation estimation as a classification task.

\subsection{Training Pose Estimation with SO(3) targets}

We describe the difficulties of training with fixed SO(3) parameterizations which will motivate the learning of view-based representations. 

\subsubsection{Regression} 

Since rotations live in a continuous space, it seems natural to directly regress a fixed SO(3) parameterizations like quaternions. However, representational constraints and pose ambiguities can introduce convergence issues as investigated by \cite{saxena2009learning}. In practice, direct regression approaches for full 3D object orientation estimation have not been very successful \citep{mahendran20173d}. Instead \cite{tremblay2018deep,tekin2017real,rad2017bb8} regress local 2D-3D correspondences and then apply a \gls{PnP} algorithm to obtain the 6D pose. However, these approaches can also not deal with pose ambiguities without additional measures (see Sec. \ref{sec:sym}).

\subsubsection{Classification} 

Classification of 3D object orientations requires a discretization of SO(3). Even rather coarse intervals of $\sim 5^o$ lead to over 50.000 possible classes. Since each class appears only sparsely in the training data, this hinders convergence. In SSD6D \citep{kehl2017ssd} the 3D orientation is learned by separately classifying a discretized viewpoint and in-plane rotation, thus reducing the complexity to $\mathcal{O}(n^2)$. However, for non-canonical views, e.g. if an object is seen from above, a change of viewpoint can be nearly equivalent to a change of in-plane rotation which yields ambiguous class combinations. In general, the relation between different orientations is ignored when performing one-hot classification.

\subsubsection{Symmetries} 
\label{sec:sym}
\begin{figure}[t]
	\centering
	\subfloat[Object symmetries\label{fig:sym}]{{\includegraphics[width=0.29\textwidth,height=0.17\textheight,keepaspectratio]{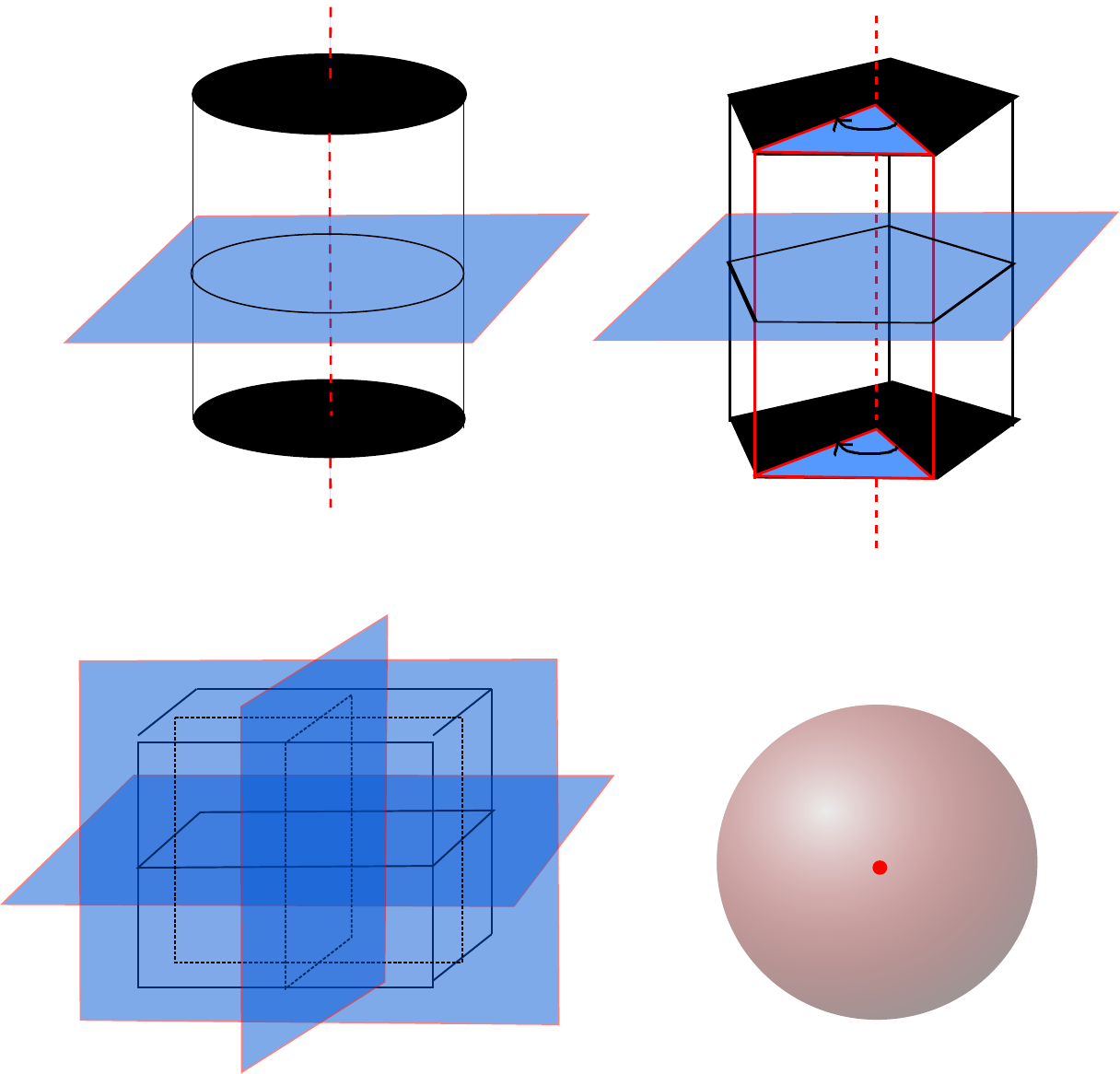} }}%
	\qquad
	\qquad
	\qquad
	\qquad
	\captionsetup{width=0.23\textwidth,,justification = raggedright}
	\subfloat[Self-occlusion \mbox{induced symmetries}\label{fig:view_sym}]{{\includegraphics[width=0.17\textwidth,height=0.15\textheight,keepaspectratio]{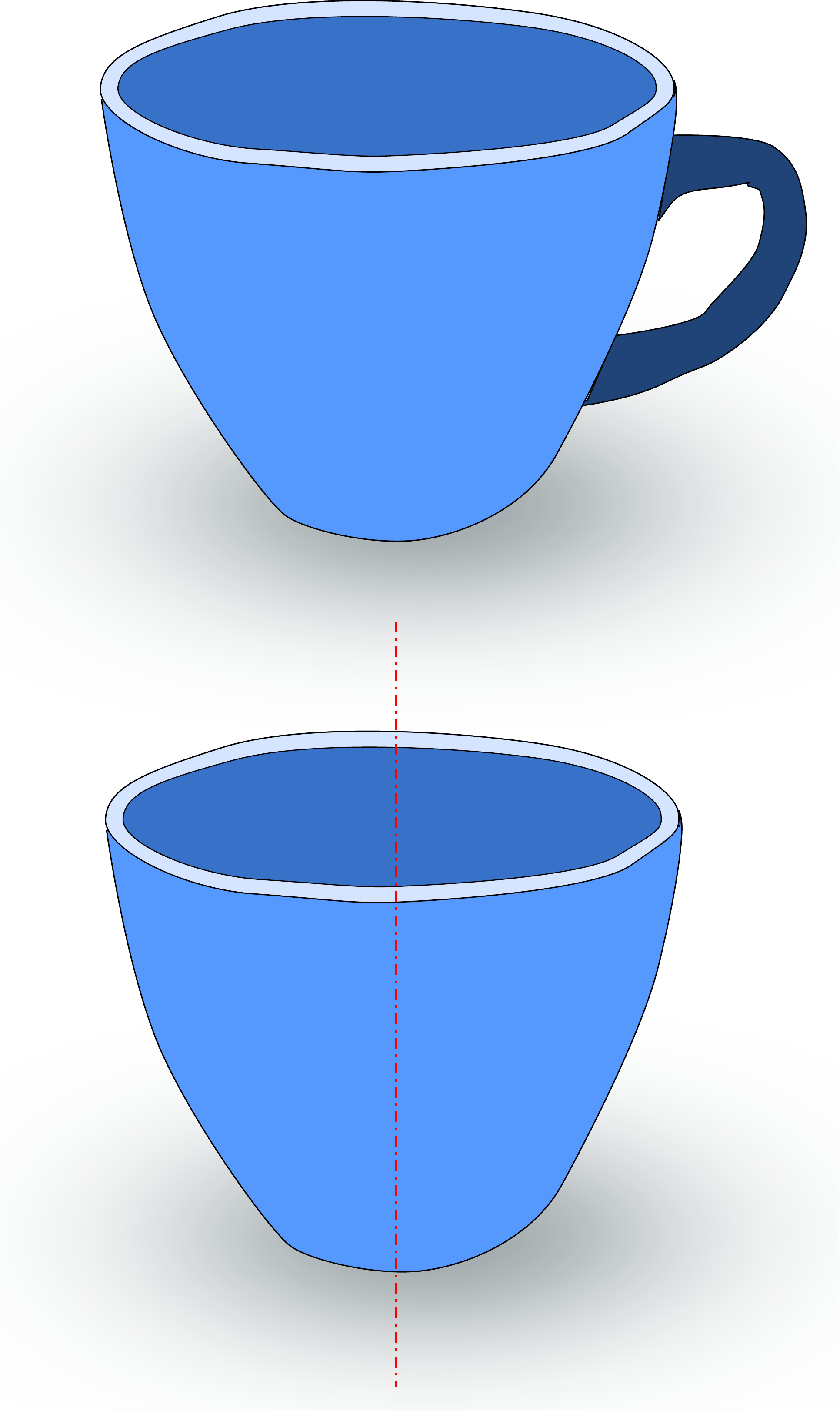} }}%
	\qquad
	\quad
	\captionsetup{width=0.205\textwidth,justification = raggedright}
	\subfloat[Occlusion \mbox{induced symmetries}\label{fig:occl}]{{\includegraphics[width=0.175\textwidth,height=0.15\textheight,keepaspectratio]{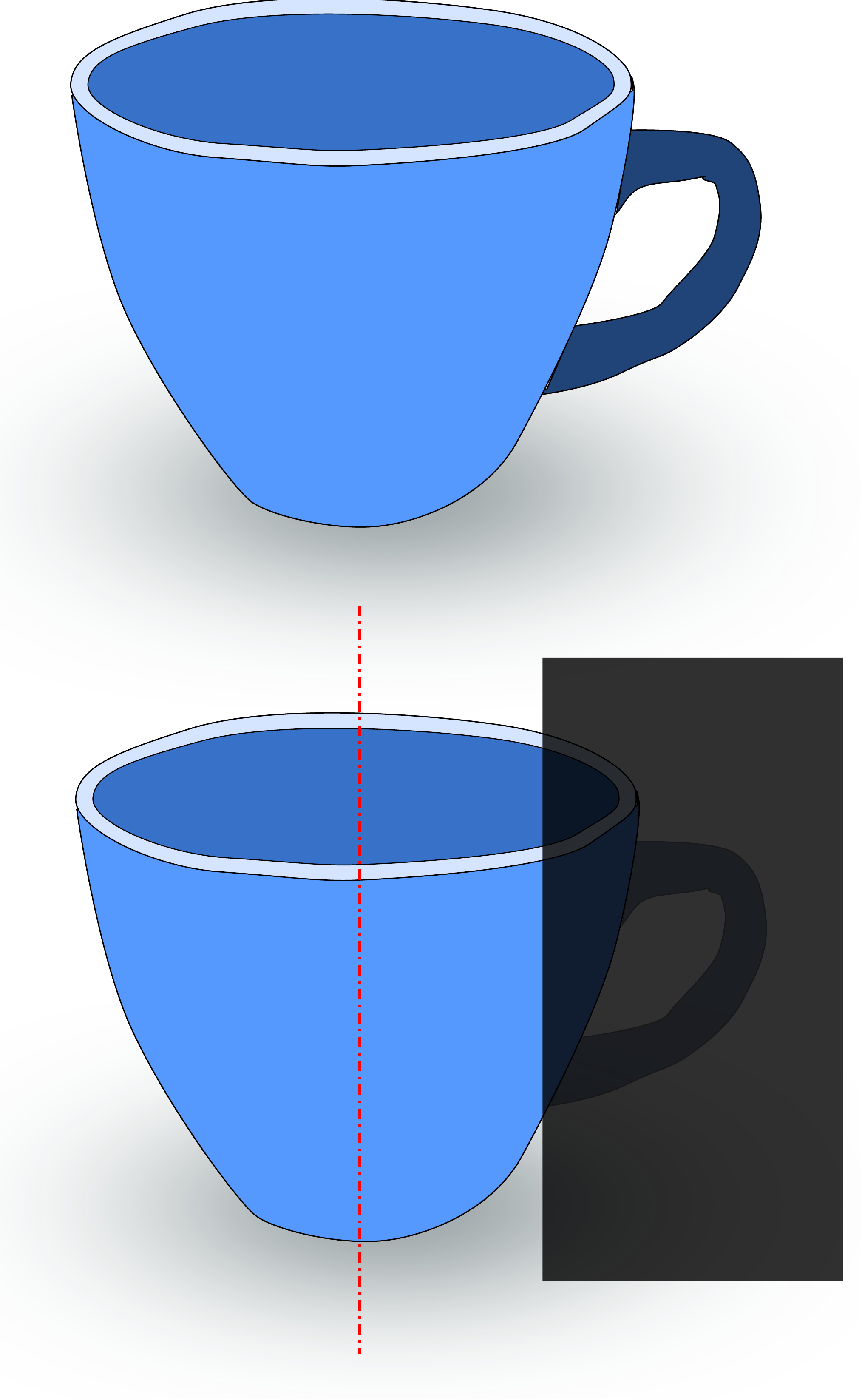} }}%
	\captionsetup{width=0.9\textwidth}
	\caption{Causes of pose ambiguities}
	\label{fig:poseamb}%
\end{figure}

Symmetries are a severe issue when relying on fixed representations of 3D orientations since they cause pose ambiguities (Fig. \ref{fig:poseamb}). If not manually addressed, identical training images can have different orientation labels assigned which can significantly disturb the learning process. In order to cope with ambiguous objects, most approaches in literature are manually adapted \citep{wohlhart2015learning,hinterstoisser2012gradient,kehl2017ssd,rad2017bb8}. The strategies reach from ignoring one axis of rotation \citep{wohlhart2015learning,hinterstoisser2012gradient} over adapting the discretization according to the object \citep{kehl2017ssd} to the training of an extra \gls{CNN} to predict symmetries \citep{rad2017bb8}. These depict tedious, manual ways to filter out object symmetries (Fig. \ref{fig:sym}) in advance, but treating ambiguities due to self-occlusions (Fig. \ref{fig:view_sym}) and occlusions (Fig. \ref{fig:occl}) are harder to address. 

Symmetries do not only affect regression and classification methods, but any learning-based algorithm that discriminates object views solely by fixed SO(3) representations.

\subsection{Learning Representations of 3D orientations}

We can also learn indirect pose representations that relate object views in a low-dimensional space. The descriptor learning can either be self-supervised by the object views themselves or still rely on fixed SO(3) representations.
\begin{figure*}[t]
	\centering
	\begin{minipage}{0.454\linewidth}
		\centering
		\captionsetup{justification=centering,font=scriptsize,aboveskip=0.15cm,belowskip=0.25cm}
		\includegraphics[width=0.74\linewidth]{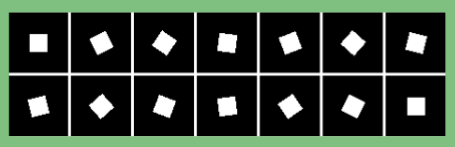}
		\caption*{\scriptsize{(a) $X_{s=1.0,t_{xy}=0.0,r \in [0,2\pi]}$}}
		\includegraphics[width=0.74\linewidth]{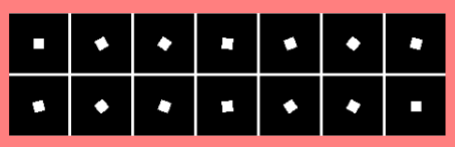}
		\caption*{\scriptsize{(b) $X_{s=0.6,t_{xy}=0.0,r \in [0,2\pi]}$}}
		\includegraphics[width=0.74\linewidth]{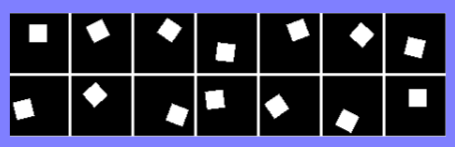}
		\caption*{\scriptsize{(c) $X_{s=1.0,t_{xy} \sim \mathcal{U}(-1,1),r \in [0,2\pi]}$}}
		\includegraphics[width=0.74\linewidth]{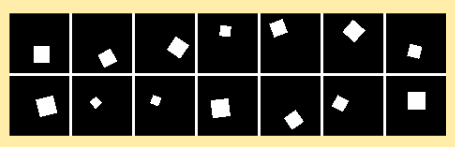}
		\caption*{\scriptsize{(d) $X_{s\sim\mathcal{U}(0.5,1),t_{xy} \sim \mathcal{U}(-1,1),r \in [0,2\pi]}$}}
	\end{minipage}%
	\begin{minipage}{0.454\linewidth}		
		\captionsetup{justification=centering,aboveskip=0.03cm,belowskip=0.12cm}
		\resizebox{1.0\linewidth}{!}{
\begin{tikzpicture}
\huge
\begin{axis}[
ylabel={z1},
ylabel={z1},
every axis y label/.style={
	at={(axis description cs:-0.15,0.5)},
	rotate=90,anchor=south
},
xmin=0, xmax=360,
ymin=-1.09470237493515, ymax=1.09969390630722,
tick align=outside,
tick pos=left,
xtick={0,90,180,270,360},
xmajorgrids,
x grid style={lightgray!92.02614379084967!black},
ymajorgrids,
y grid style={lightgray!92.02614379084967!black}
]
\addplot [ultra thick, green!50.19607843137255!black, opacity=0.7, forget plot]
table {%
0 0.994171440601349
9.23076923076923 0.99898898601532
18.4615384615385 0.779326975345612
27.6923076923077 0.961582660675049
36.9230769230769 -0.69080775976181
46.1538461538461 -0.21325720846653
55.3846153846154 0.175895154476166
64.6153846153846 0.467014998197556
73.8461538461538 0.769869267940521
83.0769230769231 0.946619391441345
92.3076923076923 0.994171440601349
101.538461538462 0.907527565956116
110.769230769231 0.859649777412415
120 0.786737561225891
129.230769230769 -0.292161792516708
138.461538461538 0.193697735667229
147.692307692308 0.237555950880051
156.923076923077 0.535309135913849
166.153846153846 0.80486649274826
175.384615384615 0.930059373378754
184.615384615385 0.914299666881561
193.846153846154 0.895997643470764
203.076923076923 0.396118879318237
212.307692307692 -0.994957089424133
221.538461538462 -0.637434720993042
230.769230769231 -0.128077253699303
240 0.424305409193039
249.230769230769 0.660383820533752
258.461538461538 0.834618449211121
267.692307692308 0.994171440601349
276.923076923077 0.997290074825287
286.153846153846 0.865739703178406
295.384615384615 0.961582660675049
304.615384615385 -0.845649003982544
313.846153846154 -0.21325720846653
323.076923076923 0.0539141893386841
332.307692307692 0.467014998197556
341.538461538461 0.692923605442047
350.769230769231 0.937107145786285
360 0.994171440601349
};
\addplot [ultra thick, red, opacity=0.7, dashed, forget plot]
table {%
0 0.91081690788269
9.23076923076923 0.958172023296356
18.4615384615385 0.946591794490814
27.6923076923077 0.96110874414444
36.9230769230769 0.94622814655304
46.1538461538461 0.882437586784363
55.3846153846154 0.796923041343689
64.6153846153846 0.753115773200989
73.8461538461538 0.851116538047791
83.0769230769231 0.90242737531662
92.3076923076923 0.91081690788269
101.538461538462 0.926050901412964
110.769230769231 0.953774213790894
120 0.958892166614532
129.230769230769 0.945202648639679
138.461538461538 0.859587132930756
147.692307692308 0.769328951835632
156.923076923077 0.752761542797089
166.153846153846 0.8467777967453
175.384615384615 0.884642481803894
184.615384615385 0.997358798980713
193.846153846154 0.921492218971252
203.076923076923 0.952597677707672
212.307692307692 0.961271524429321
221.538461538462 0.922319710254669
230.769230769231 0.810805380344391
240 0.769846022129059
249.230769230769 0.788207352161407
258.461538461538 0.847922682762146
267.692307692308 0.91081690788269
276.923076923077 0.994028508663177
286.153846153846 0.917502343654633
295.384615384615 0.964515686035156
304.615384615385 0.940130174160004
313.846153846154 0.880700886249542
323.076923076923 0.841343641281128
332.307692307692 0.77567332983017
341.538461538461 0.799918949604034
350.769230769231 0.915483891963959
360 0.91081690788269
};
\addplot [ultra thick, blue, opacity=0.7, dashed, forget plot]
table {%
0 0.556879460811615
9.23076923076923 0.662064611911774
18.4615384615385 0.258758753538132
27.6923076923077 0.570461571216583
36.9230769230769 0.723670482635498
46.1538461538461 0.261409789323807
55.3846153846154 0.801155865192413
64.6153846153846 0.978064894676208
73.8461538461538 0.834719955921173
83.0769230769231 0.191523522138596
92.3076923076923 0.161711618304253
101.538461538462 0.254533231258392
110.769230769231 0.9043830037117
120 0.999548077583313
129.230769230769 0.999948620796204
138.461538461538 0.995857656002045
147.692307692308 0.337354183197021
156.923076923077 0.441515207290649
166.153846153846 0.540768563747406
175.384615384615 0.567718327045441
184.615384615385 0.542390525341034
193.846153846154 0.751091599464417
203.076923076923 0.681029915809631
212.307692307692 0.588562488555908
221.538461538462 0.813160836696625
230.769230769231 0.982178390026093
240 0.906164824962616
249.230769230769 0.231224954128265
258.461538461538 0.320656687021255
267.692307692308 0.99797523021698
276.923076923077 0.915059983730316
286.153846153846 0.0422875545918941
295.384615384615 0.856416702270508
304.615384615385 0.521307647228241
313.846153846154 0.473534256219864
323.076923076923 0.910015881061554
332.307692307692 0.555698215961456
341.538461538461 0.997273564338684
350.769230769231 0.314454793930054
360 0.55700558423996
};
\end{axis}

\end{tikzpicture}%
\begin{tikzpicture}
\huge
\begin{axis}[
ylabel={z1},
ylabel={z2},
every axis y label/.style={
	at={(axis description cs:-0.15,0.5)},
	rotate=90,anchor=south
},
xmin=0, xmax=360,
ymin=-1.09470237493515, ymax=1.09969390630722,
tick align=outside,
tick pos=left,
xtick={0,90,180,270,360},
xmajorgrids,
x grid style={lightgray!92.02614379084967!black},
ymajorgrids,
y grid style={lightgray!92.02614379084967!black}
]
\addplot [ultra thick, green!50.19607843137255!black, opacity=0.7, forget plot]
table {%
0 -0.107810899615288
9.23076923076923 0.0449541099369526
18.4615384615385 0.626617550849915
27.6923076923077 0.274515509605408
36.9230769230769 -0.7230384349823
46.1538461538461 -0.976996064186096
55.3846153846154 -0.984408855438232
64.6153846153846 -0.8842493891716
73.8461538461538 -0.638201653957367
83.0769230769231 -0.322353541851044
92.3076923076923 -0.107810899615288
101.538461538462 0.419992506504059
110.769230769231 0.510883748531342
120 -0.617287635803223
129.230769230769 -0.956368923187256
138.461538461538 -0.981061279773712
147.692307692308 -0.971373856067657
156.923076923077 -0.84465616941452
166.153846153846 -0.593455910682678
175.384615384615 -0.367409348487854
184.615384615385 0.405038625001907
193.846153846154 0.444058746099472
203.076923076923 0.918199241161346
212.307692307692 -0.100301317870617
221.538461538462 -0.770504355430603
230.769230769231 -0.991764187812805
240 -0.905519187450409
249.230769230769 -0.750928223133087
258.461538461538 -0.550828516483307
267.692307692308 -0.107810899615288
276.923076923077 0.0735695213079453
286.153846153846 0.500494539737701
295.384615384615 0.274515509605408
304.615384615385 -0.533739328384399
313.846153846154 -0.976996064186096
323.076923076923 -0.998545587062836
332.307692307692 -0.8842493891716
341.538461538461 -0.721010982990265
350.769230769231 -0.349042028188705
360 -0.107810899615288
};
\addplot [ultra thick, red, opacity=0.7, dashed, forget plot]
table {%
0 -0.412810415029526
9.23076923076923 -0.286192804574966
18.4615384615385 -0.322434484958649
27.6923076923077 -0.276170194149017
36.9230769230769 -0.323500096797943
46.1538461538461 -0.470429539680481
55.3846153846154 -0.60408079624176
64.6153846153846 -0.657887995243073
73.8461538461538 -0.524976849555969
83.0769230769231 -0.430842041969299
92.3076923076923 -0.412810415029526
101.538461538462 -0.377398818731308
110.769230769231 -0.300524175167084
120 -0.283770561218262
129.230769230769 -0.326484382152557
138.461538461538 -0.510989189147949
147.692307692308 -0.638852834701538
156.923076923077 -0.65829336643219
166.153846153846 -0.53194671869278
175.384615384615 -0.46627002954483
184.615384615385 -0.0726315602660179
193.846153846154 -0.388396829366684
203.076923076923 -0.304232835769653
212.307692307692 -0.275602966547012
221.538461538462 -0.386427789926529
230.769230769231 -0.585315823554993
240 -0.638229608535767
249.230769230769 -0.615409791469574
258.461538461538 -0.530119955539703
267.692307692308 -0.412810415029526
276.923076923077 -0.109120637178421
286.153846153846 -0.397730499505997
295.384615384615 -0.264025717973709
304.615384615385 -0.340815663337708
313.846153846154 -0.473672807216644
323.076923076923 -0.540500581264496
332.307692307692 -0.631134629249573
341.538461538461 -0.600108027458191
350.769230769231 -0.402354717254639
360 -0.412810415029526
};
\addplot [ultra thick, blue, opacity=0.7, dashed, forget plot]
table {%
0 -0.830593407154083
9.23076923076923 -0.74944669008255
18.4615384615385 -0.965941965579987
27.6923076923077 -0.821324288845062
36.9230769230769 -0.690145671367645
46.1538461538461 -0.965227901935577
55.3846153846154 -0.598455727100372
64.6153846153846 -0.208300545811653
73.8461538461538 -0.550674617290497
83.0769230769231 -0.981487989425659
92.3076923076923 -0.986838102340698
101.538461538462 -0.967063963413239
110.769230769231 -0.426721721887589
120 -0.0300623849034309
129.230769230769 -0.0101397652179003
138.461538461538 0.090926006436348
147.692307692308 -0.941377818584442
156.923076923077 -0.897253811359406
166.153846153846 -0.841171383857727
175.384615384615 -0.823222935199738
184.615384615385 -0.84012645483017
193.846153846154 -0.660197973251343
203.076923076923 -0.732255578041077
212.307692307692 -0.808451771736145
221.538461538462 -0.582038938999176
230.769230769231 -0.187950909137726
240 -0.422924697399139
249.230769230769 -0.97290027141571
258.461538461538 -0.947195529937744
267.692307692308 0.0636040344834328
276.923076923077 -0.40331768989563
286.153846153846 -0.999105453491211
295.384615384615 -0.516285240650177
304.615384615385 -0.853368759155273
313.846153846154 -0.880775392055511
323.076923076923 -0.414573460817337
332.307692307692 -0.831384062767029
341.538461538461 -0.0737925693392754
350.769230769231 -0.949272453784943
360 -0.830508708953857
};
\end{axis}

\end{tikzpicture}}
		\caption*{\scriptsize{(1) Autoencoder $\colorbox{darkgreen}{(a)}\longrightarrow \colorbox{darkgreen}{(a)}$ }}
		\resizebox{1.0\linewidth}{!}{
\begin{tikzpicture}
\huge
\begin{axis}[
ylabel={z1},
every axis y label/.style={
	at={(axis description cs:-0.15,0.5)},
	rotate=90,anchor=south
},
xmin=0, xmax=360,
ymin=-1.09470237493515, ymax=1.09969390630722,
tick align=outside,
tick pos=left,
xtick={0,90,180,270,360},
xmajorgrids,
x grid style={lightgray!92.02614379084967!black},
ymajorgrids,
y grid style={lightgray!92.02614379084967!black}
]
\addplot [ultra thick, green!50.19607843137255!black, opacity=0.7, forget plot]
table {%
0 -0.993383407592773
9.23076923076923 -0.666235327720642
18.4615384615385 -0.258563578128815
27.6923076923077 -0.0948017612099648
36.9230769230769 -0.0345274806022644
46.1538461538461 -0.132052272558212
55.3846153846154 -0.212084710597992
64.6153846153846 -0.34804230928421
73.8461538461538 -0.39956983923912
83.0769230769231 -0.884653627872467
92.3076923076923 -0.993383407592773
101.538461538462 -0.327665954828262
110.769230769231 -0.274191230535507
120 -0.118423752486706
129.230769230769 -0.0560745075345039
138.461538461538 -0.104383982717991
147.692307692308 -0.219025656580925
156.923076923077 -0.255235135555267
166.153846153846 -0.63641232252121
175.384615384615 -0.933544337749481
184.615384615385 -0.965695977210999
193.846153846154 -0.403659850358963
203.076923076923 -0.107606798410416
212.307692307692 0.0487858280539513
221.538461538462 -0.0431058257818222
230.769230769231 -0.108041197061539
240 -0.403236359357834
249.230769230769 -0.400935888290405
258.461538461538 -0.727642297744751
267.692307692308 -0.993383407592773
276.923076923077 -0.828488171100616
286.153846153846 -0.307363867759705
295.384615384615 -0.0948017612099648
304.615384615385 0.0247897803783417
313.846153846154 -0.132052272558212
323.076923076923 -0.0862478241324425
332.307692307692 -0.34804230928421
341.538461538461 -0.471784085035324
350.769230769231 -0.824225544929504
360 -0.993383407592773
};
\addplot [ultra thick, red, opacity=0.7, dashed, forget plot]
table {%
0 -0.686246514320374
9.23076923076923 -0.695610284805298
18.4615384615385 -0.664615452289581
27.6923076923077 -0.6628258228302
36.9230769230769 -0.685925424098969
46.1538461538461 -0.650201737880707
55.3846153846154 -0.669927954673767
64.6153846153846 -0.670828759670258
73.8461538461538 -0.662597477436066
83.0769230769231 -0.689926147460938
92.3076923076923 -0.686246514320374
101.538461538462 -0.672750890254974
110.769230769231 -0.659129500389099
120 -0.663612484931946
129.230769230769 -0.675131261348724
138.461538461538 -0.669033348560333
147.692307692308 -0.66722172498703
156.923076923077 -0.664245963096619
166.153846153846 -0.666699767112732
175.384615384615 -0.689550697803497
184.615384615385 -0.695738554000854
193.846153846154 -0.670084178447723
203.076923076923 -0.647148728370667
212.307692307692 -0.661815524101257
221.538461538462 -0.669012308120728
230.769230769231 -0.674086809158325
240 -0.670521318912506
249.230769230769 -0.66930741071701
258.461538461538 -0.662907004356384
267.692307692308 -0.686246514320374
276.923076923077 -0.694571554660797
286.153846153846 -0.665877282619476
295.384615384615 -0.646741330623627
304.615384615385 -0.67492401599884
313.846153846154 -0.642340958118439
323.076923076923 -0.684121489524841
332.307692307692 -0.679565906524658
341.538461538461 -0.660180985927582
350.769230769231 -0.681871175765991
360 -0.686246514320374
};
\addplot [ultra thick, blue, opacity=0.7, dashed, forget plot]
table {%
0 -0.770898103713989
9.23076923076923 0.00182325206696987
18.4615384615385 -0.34413954615593
27.6923076923077 -0.365553200244904
36.9230769230769 -0.868417620658875
46.1538461538461 -0.882582247257233
55.3846153846154 0.993793368339539
64.6153846153846 -0.193689420819283
73.8461538461538 -0.708338439464569
83.0769230769231 0.276629626750946
92.3076923076923 0.637715816497803
101.538461538462 -0.0737057253718376
110.769230769231 -0.976689696311951
120 -0.141309380531311
129.230769230769 -0.816475749015808
138.461538461538 0.00698873167857528
147.692307692308 -0.863076329231262
156.923076923077 0.929317772388458
166.153846153846 -0.102158553898335
175.384615384615 -0.214593082666397
184.615384615385 0.143456861376762
193.846153846154 -0.164940789341927
203.076923076923 0.882796227931976
212.307692307692 -0.810533761978149
221.538461538462 -0.198935598134995
230.769230769231 -0.233030959963799
240 -0.243293464183807
249.230769230769 -0.698858499526978
258.461538461538 -0.988217294216156
267.692307692308 0.597466945648193
276.923076923077 0.775664865970612
286.153846153846 0.0794311538338661
295.384615384615 -0.194250777363777
304.615384615385 0.778901815414429
313.846153846154 0.914871573448181
323.076923076923 -0.352955371141434
332.307692307692 -0.328127205371857
341.538461538461 -0.357623070478439
350.769230769231 0.993402659893036
360 -0.970343887805939
};
\end{axis}

\end{tikzpicture}%
\begin{tikzpicture}
\huge
\begin{axis}[
ylabel={z2},
every axis y label/.style={
	at={(axis description cs:-0.15,0.5)},
	rotate=90,anchor=south
},
xmin=0, xmax=360,
ymin=-1.09470237493515, ymax=1.09969390630722,
tick align=outside,
tick pos=left,
xtick={0,90,180,270,360},
xmajorgrids,
x grid style={lightgray!92.02614379084967!black},
ymajorgrids,
y grid style={lightgray!92.02614379084967!black}
]
\addplot [ultra thick, green!50.19607843137255!black, opacity=0.7, forget plot]
table {%
0 -0.114844754338264
9.23076923076923 -0.745741546154022
18.4615384615385 -0.965994238853455
27.6923076923077 -0.995496213436127
36.9230769230769 -0.999403774738312
46.1538461538461 -0.99124276638031
55.3846153846154 -0.977251291275024
64.6153846153846 -0.937478840351105
73.8461538461538 -0.916702806949615
83.0769230769231 -0.466248840093613
92.3076923076923 -0.114844754338264
101.538461538462 -0.94479364156723
110.769230769231 -0.96167516708374
120 -0.992963194847107
129.230769230769 -0.998426616191864
138.461538461538 -0.994537055492401
147.692307692308 -0.975719153881073
156.923076923077 -0.9668790102005
166.153846153846 -0.77134907245636
175.384615384615 -0.358461856842041
184.615384615385 -0.259675443172455
193.846153846154 -0.91490912437439
203.076923076923 -0.994193494319916
212.307692307692 -0.998809278011322
221.538461538462 -0.999070525169373
230.769230769231 -0.994146406650543
240 -0.915095865726471
249.230769230769 -0.916106164455414
258.461538461538 -0.685956716537476
267.692307692308 -0.114844754338264
276.923076923077 -0.560006499290466
286.153846153846 -0.951592147350311
295.384615384615 -0.995496213436127
304.615384615385 -0.999692678451538
313.846153846154 -0.99124276638031
323.076923076923 -0.996273756027222
332.307692307692 -0.937478840351105
341.538461538461 -0.881714105606079
350.769230769231 -0.56626158952713
360 -0.114844754338264
};
\addplot [ultra thick, red, opacity=0.7, dashed, forget plot]
table {%
0 -0.727369070053101
9.23076923076923 -0.718419373035431
18.4615384615385 -0.747185587882996
27.6923076923077 -0.748773574829102
36.9230769230769 -0.727671802043915
46.1538461538461 -0.7597616314888
55.3846153846154 -0.742426037788391
64.6153846153846 -0.741612255573273
73.8461538461538 -0.74897563457489
83.0769230769231 -0.72387969493866
92.3076923076923 -0.727369070053101
101.538461538462 -0.739869117736816
110.769230769231 -0.752029418945312
120 -0.748076558113098
129.230769230769 -0.737697541713715
138.461538461538 -0.743232369422913
147.692307692308 -0.744859099388123
156.923076923077 -0.747514069080353
166.153846153846 -0.745326399803162
175.384615384615 -0.724237501621246
184.615384615385 -0.718295156955719
193.846153846154 -0.742285132408142
203.076923076923 -0.762363851070404
212.307692307692 -0.749666690826416
221.538461538462 -0.743251323699951
230.769230769231 -0.738652050495148
240 -0.741890251636505
249.230769230769 -0.742985606193542
258.461538461538 -0.748701691627502
267.692307692308 -0.727369070053101
276.923076923077 -0.719423532485962
286.153846153846 -0.746061265468597
295.384615384615 -0.762709438800812
304.615384615385 -0.73788720369339
313.846153846154 -0.766418993473053
323.076923076923 -0.729368090629578
332.307692307692 -0.733614444732666
341.538461538461 -0.751106560230255
350.769230769231 -0.731472313404083
360 -0.727369070053101
};
\addplot [ultra thick, blue, opacity=0.7, dashed, forget plot]
table {%
0 -0.636958479881287
9.23076923076923 0.999998331069946
18.4615384615385 0.938918471336365
27.6923076923077 -0.930790424346924
36.9230769230769 0.495833426713943
46.1538461538461 0.47015792131424
55.3846153846154 -0.111241795122623
64.6153846153846 -0.981062948703766
73.8461538461538 0.70587295293808
83.0769230769231 0.960976660251617
92.3076923076923 -0.770271718502045
101.538461538462 0.99728000164032
110.769230769231 -0.21465602517128
120 -0.989965438842773
129.230769230769 -0.577379703521729
138.461538461538 -0.999975621700287
147.692307692308 -0.505073547363281
156.923076923077 -0.369281113147736
166.153846153846 -0.994768142700195
175.384615384615 -0.976703464984894
184.615384615385 0.989656567573547
193.846153846154 -0.986303508281708
203.076923076923 -0.469756096601486
212.307692307692 0.585691928863525
221.538461538462 -0.980012536048889
230.769230769231 -0.972469329833984
240 0.969952762126923
249.230769230769 0.715259969234467
258.461538461538 -0.153057605028152
267.692307692308 0.801893532276154
276.923076923077 -0.631144881248474
286.153846153846 0.996840357780457
295.384615384615 -0.980951964855194
304.615384615385 -0.627145886421204
313.846153846154 0.403744965791702
323.076923076923 0.935640156269073
332.307692307692 0.944633543491364
341.538461538461 -0.933866024017334
350.769230769231 -0.114678479731083
360 0.241728648543358
};
\end{axis}

\end{tikzpicture}}
		\caption*{\scriptsize{(2) Autoencoder $\colorbox{darkyellow}{(d)} \longrightarrow \colorbox{darkyellow}{(d)}$}}
		\resizebox{1.0\linewidth}{!}{
\begin{tikzpicture}
\huge
\begin{axis}[
xlabel={rotation angle [deg]},
ylabel={z1},
every axis x label/.style={
	at={(axis description cs:0.5,-0.15)},
	anchor=north,
},
every axis y label/.style={
	at={(axis description cs:-0.15,0.5)},
	rotate=90,anchor=south
},
xmin=0, xmax=360,
ymin=-1.09470237493515, ymax=1.09969390630722,
tick align=outside,
tick pos=left,
xtick={0,90,180,270,360},
xmajorgrids,
x grid style={lightgray!92.02614379084967!black},
ymajorgrids,
y grid style={lightgray!92.02614379084967!black},
]
\addplot [ultra thick, green!50.19607843137255!black, opacity=0.7]
table {%
0 -0.870365262031555
9.23076923076923 -0.975093305110931
18.4615384615385 -0.479809582233429
27.6923076923077 0.0510527528822422
36.9230769230769 0.654303729534149
46.1538461538461 0.92222648859024
55.3846153846154 0.983726143836975
64.6153846153846 0.831017136573792
73.8461538461538 0.29422852396965
83.0769230769231 -0.425079733133316
92.3076923076923 -0.870365262031555
101.538461538462 -0.796322464942932
110.769230769231 -0.324108719825745
120 0.237956970930099
129.230769230769 0.759101331233978
138.461538461538 0.991937696933746
147.692307692308 0.94775402545929
156.923076923077 0.701335728168488
166.153846153846 0.123516008257866
175.384615384615 -0.567686796188354
184.615384615385 -0.999692916870117
193.846153846154 -0.721905529499054
203.076923076923 -0.182790726423264
212.307692307692 0.345999658107758
221.538461538462 0.788688480854034
230.769230769231 0.997153401374817
240 0.914795696735382
249.230769230769 0.599163830280304
258.461538461538 -0.0400488339364529
267.692307692308 -0.870365262031555
276.923076923077 -0.975455462932587
286.153846153846 -0.622779726982117
295.384615384615 0.0510527528822422
304.615384615385 0.51299911737442
313.846153846154 0.92222648859024
323.076923076923 0.999085545539856
332.307692307692 0.831017136573792
341.538461538461 0.459094047546387
350.769230769231 -0.438419133424759
360 -0.870365262031555
};
\addplot [ultra thick, red, opacity=0.7, dashed]
table {%
0 -0.79953545331955
9.23076923076923 -0.831400811672211
18.4615384615385 -0.448578178882599
27.6923076923077 0.268102556467056
36.9230769230769 0.707609593868256
46.1538461538461 0.766650080680847
55.3846153846154 0.888265192508698
64.6153846153846 0.74179607629776
73.8461538461538 0.176251500844955
83.0769230769231 -0.623833477497101
92.3076923076923 -0.869403004646301
101.538461538462 -0.695741772651672
110.769230769231 -0.277245849370956
120 0.293956160545349
129.230769230769 0.665268123149872
138.461538461538 0.804492473602295
147.692307692308 0.893734574317932
156.923076923077 0.700767815113068
166.153846153846 0.141468465328217
175.384615384615 -0.574940860271454
184.615384615385 -0.929569065570831
193.846153846154 -0.59324324131012
203.076923076923 -0.168348804116249
212.307692307692 0.326462209224701
221.538461538462 0.618596196174622
230.769230769231 0.875596940517426
240 0.867485284805298
249.230769230769 0.568007588386536
258.461538461538 -0.0162110347300768
267.692307692308 -0.822062373161316
276.923076923077 -0.992038011550903
286.153846153846 -0.591395318508148
295.384615384615 -0.104116566479206
304.615384615385 0.477032542228699
313.846153846154 0.725046753883362
323.076923076923 0.941367566585541
332.307692307692 0.910088896751404
341.538461538461 0.461303532123566
350.769230769231 -0.381045550107956
360 -0.734779000282288
};
\addplot [ultra thick, blue, opacity=0.7, dashed]
table {%
0 -0.912014365196228
9.23076923076923 -0.932088017463684
18.4615384615385 -0.486162006855011
27.6923076923077 0.156060323119164
36.9230769230769 0.622016131877899
46.1538461538461 0.957151293754578
55.3846153846154 0.980969309806824
64.6153846153846 0.81073796749115
73.8461538461538 0.29637348651886
83.0769230769231 -0.53644186258316
92.3076923076923 -0.952829301357269
101.538461538462 -0.863821983337402
110.769230769231 -0.318656653165817
120 0.237155690789223
129.230769230769 0.749033391475677
138.461538461538 0.980664670467377
147.692307692308 0.955034613609314
156.923076923077 0.703853130340576
166.153846153846 0.0400237068533897
175.384615384615 -0.703634262084961
184.615384615385 -0.993806660175323
193.846153846154 -0.756892323493958
203.076923076923 -0.131376937031746
212.307692307692 0.373752892017365
221.538461538462 0.827586650848389
230.769230769231 0.996175825595856
240 0.935174703598022
249.230769230769 0.586285710334778
258.461538461538 -0.09526526927948
267.692307692308 -0.862291812896729
276.923076923077 -0.99026083946228
286.153846153846 -0.629620254039764
295.384615384615 -0.0127649279311299
304.615384615385 0.51507180929184
313.846153846154 0.91376930475235
323.076923076923 0.99944531917572
332.307692307692 0.878856539726257
341.538461538461 0.458964139223099
350.769230769231 -0.428029030561447
360 -0.899776101112366
};
\end{axis}

\end{tikzpicture}%
\begin{tikzpicture}
\huge
\begin{axis}[
xlabel={rotation angle [deg]},
ylabel={z2},
every axis x label/.style={
	at={(axis description cs:0.5,-0.15)},
	anchor=north,
},
every axis y label/.style={
	at={(axis description cs:-0.15,0.5)},
	rotate=90,anchor=south
},
xmin=0, xmax=360,
ymin=-1.09470237493515, ymax=1.09969390630722,
tick align=outside,
tick pos=left,
xtick={0,90,180,270,360},
xmajorgrids,
x grid style={lightgray!92.02614379084967!black},
ymajorgrids,
y grid style={lightgray!92.02614379084967!black},
]
\addplot [ultra thick, green!50.19607843137255!black, opacity=0.7]
table {%
0 0.492406636476517
9.23076923076923 -0.221794947981834
18.4615384615385 -0.877372682094574
27.6923076923077 -0.998695909976959
36.9230769230769 -0.756231904029846
46.1538461538461 -0.386650294065475
55.3846153846154 0.179674103856087
64.6153846153846 0.556246817111969
73.8461538461538 0.955735087394714
83.0769230769231 0.905155897140503
92.3076923076923 0.492406636476517
101.538461538462 -0.604872286319733
110.769230769231 -0.946019887924194
120 -0.971275746822357
129.230769230769 -0.650972545146942
138.461538461538 -0.126726314425468
147.692307692308 0.319001853466034
156.923076923077 0.712831139564514
166.153846153846 0.992342591285706
175.384615384615 0.823244631290436
184.615384615385 -0.0247820541262627
193.846153846154 -0.691991567611694
203.076923076923 -0.983151853084564
212.307692307692 -0.938234567642212
221.538461538462 -0.614793062210083
230.769230769231 -0.075399674475193
240 0.403916716575623
249.230769230769 0.800626456737518
258.461538461538 0.999197781085968
267.692307692308 0.492406636476517
276.923076923077 -0.220196917653084
286.153846153846 -0.782397210597992
295.384615384615 -0.998695909976959
304.615384615385 -0.85838919878006
313.846153846154 -0.386650294065475
323.076923076923 0.0427547320723534
332.307692307692 0.556246817111969
341.538461538461 0.888387739658356
350.769230769231 0.898770689964294
360 0.492406636476517
};
\addplot [ultra thick, red, opacity=0.7, dashed]
table {%
0 0.389654338359833
9.23076923076923 -0.253565013408661
18.4615384615385 -0.824757099151611
27.6923076923077 -0.990434169769287
36.9230769230769 -0.665586769580841
46.1538461538461 -0.286244988441467
55.3846153846154 0.175207003951073
64.6153846153846 0.650999784469604
73.8461538461538 0.900573492050171
83.0769230769231 0.931883633136749
92.3076923076923 0.423704355955124
101.538461538462 -0.527175962924957
110.769230769231 -0.885324001312256
120 -0.878199875354767
129.230769230769 -0.628830015659332
138.461538461538 -0.0691161006689072
147.692307692308 0.320285379886627
156.923076923077 0.697411179542542
166.153846153846 0.988571465015411
175.384615384615 0.750863969326019
184.615384615385 -0.0535352304577827
193.846153846154 -0.6277015209198
203.076923076923 -1.00186777114868
212.307692307692 -0.874188601970673
221.538461538462 -0.565432786941528
230.769230769231 -0.0544717982411385
240 0.340980887413025
249.230769230769 0.738310754299164
258.461538461538 0.841007232666016
267.692307692308 0.400632858276367
276.923076923077 -0.118898078799248
286.153846153846 -0.683422684669495
295.384615384615 -0.995773077011108
304.615384615385 -0.802426755428314
313.846153846154 -0.355017066001892
323.076923076923 -0.0539322756230831
332.307692307692 0.403337627649307
341.538461538461 0.92145174741745
350.769230769231 0.849914252758026
360 0.358095198869705
};
\addplot [ultra thick, blue, opacity=0.7, dashed]
table {%
0 0.410158336162567
9.23076923076923 -0.362231999635696
18.4615384615385 -0.873868703842163
27.6923076923077 -0.987747490406036
36.9230769230769 -0.783004403114319
46.1538461538461 -0.289588272571564
55.3846153846154 0.194163054227829
64.6153846153846 0.585409224033356
73.8461538461538 0.955072104930878
83.0769230769231 0.843937277793884
92.3076923076923 0.30350661277771
101.538461538462 -0.503797173500061
110.769230769231 -0.947870194911957
120 -0.971471607685089
129.230769230769 -0.662532269954681
138.461538461538 -0.195695549249649
147.692307692308 0.296494483947754
156.923076923077 0.710345566272736
166.153846153846 0.999198734760284
175.384615384615 0.710562229156494
184.615384615385 0.111122764647007
193.846153846154 -0.653539538383484
203.076923076923 -0.991332471370697
212.307692307692 -0.927528262138367
221.538461538462 -0.561338007450104
230.769230769231 -0.087370902299881
240 0.354186713695526
249.230769230769 0.810104370117188
258.461538461538 0.995451867580414
267.692307692308 0.506411671638489
276.923076923077 -0.139224678277969
286.153846153846 -0.776903033256531
295.384615384615 -0.999918520450592
304.615384615385 -0.857147037982941
313.846153846154 -0.406233578920364
323.076923076923 0.0333028063178062
332.307692307692 0.477086186408997
341.538461538461 0.888454854488373
350.769230769231 0.903765022754669
360 0.436351835727692
};
\end{axis}

\end{tikzpicture}}
		\caption*{\scriptsize{(3) Augmented Autoencoder $\colorbox{darkyellow}{(d)} \longrightarrow \colorbox{darkgreen}{(a)}$}}
	\end{minipage}
	\caption{Experiment on the dsprites dataset of \cite{dsprites17}. Left: 64x64 squares from four distributions (a,b,c and d) distinguished by \mbox{scale ($s$)} and translation ($t_{xy}$) that are used for training and testing. Right: Normalized latent dimensions $z_1$ and $z_2$ for all rotations ($r$) of the distribution (a), (b) or (c) after training ordinary \glspl{AE} (1),(2) and an \gls{AAE} (3) to reconstruct squares of the same orientation.}
	\label{fig:toy}
\end{figure*}

\subsubsection{Descriptor Learning}

\cite{wohlhart2015learning} introduced a \gls{CNN}-based descriptor learning approach using a triplet loss that minimizes/maximizes the Euclidean distance between similar/dissimilar object orientations. In addition, the distance between different objects is maximized.
Although mixing in synthetic data, the training also relies on pose-annotated sensor data. The approach is not immune against symmetries since the descriptor is built using explicit 3D orientations. Thus, the loss can be dominated by symmetric object views that appear the same but have opposite orientations which can produce incorrect average pose predictions.

\cite{balntas2017pose} extended this work by enforcing proportionality between descriptor and pose distances. They acknowledge the problem of object symmetries by weighting the pose distance loss with the depth difference of the object at the considered poses. This heuristic increases the accuracy on symmetric objects with respect to \cite{wohlhart2015learning}. 

Our work is also based on learning descriptors, but in contrast we train our Augmented Autoencoders (AAEs) such that the learning process itself is independent of any fixed SO(3) representation. The loss is solely based on the appearance of the reconstructed object views and thus symmetrical ambiguities are inherently regarded. Thus, unlike \cite{balntas2017pose,wohlhart2015learning} we abstain from the use of real labeled data during training and instead train completely self-supervised. This means that assigning 3D orientations to the descriptors only happens after the training.

\cite{kehl2016deep} train an Autoencoder architecture on random RGB-D scene patches from the LineMOD dataset \cite{hinterstoisser2011multimodal}. At test time, descriptors from scene and object patches are compared to find the 6D pose.
Since the approach requires the evaluation of a lot of patches, it takes about 670ms per prediction. Furthermore, using local patches means to ignore holistic relations between object features which is crucial if few texture exists. Instead we train on holistic object views and explicitly learn domain invariance.

\section{Method}

In the following, we mainly focus on the novel 3D orientation estimation technique based on the AAE.
\begin{figure*}[t]
	\centering
	\captionsetup{width=0.9\textwidth}
	\includegraphics[width=0.8\textwidth]{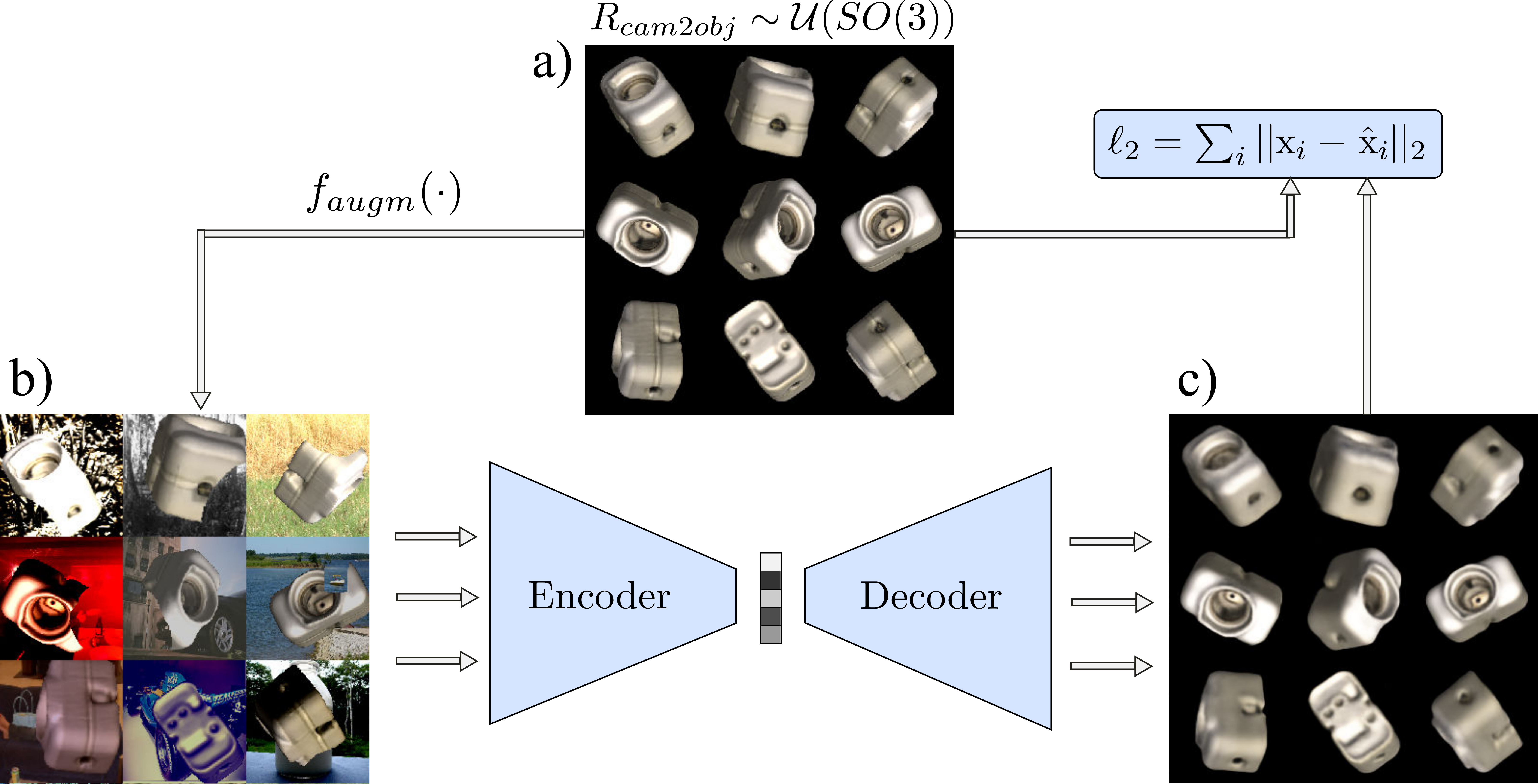}
	\caption{Training process for the \gls{AAE}; a) reconstruction target batch $\pmb x$ of uniformly sampled SO(3) object views; b) geometric and color augmented input; c) reconstruction $\pmb{\hat{x}}$ after 40000 iterations}
	\label{fig:training_process}
\end{figure*}
\subsection{Autoencoders}

The original \gls{AE}, introduced by \cite{rumelhart1985learning}, is a dimensionality reduction technique for high dimensional data such as images, audio or depth. It consists of an Encoder $\Phi$ and a Decoder $\Psi$, both arbitrary learnable function approximators which are usually neural networks. The training objective is to reconstruct the input $x \in \mathcal{R}^{\mathcal{D}}$ after passing through a low-dimensional bottleneck, referred to as the latent representation $z \in \mathcal{R}^{n}$ with $n << \mathcal{D}$ :
\begin{equation}
\label{eq:yhat}
\hat{x} = (\Psi\circ\Phi)(x) = \Psi(z)
\end{equation}
The per-sample loss is simply a sum over the pixel-wise L2 distance
\begin{equation}
\label{eq:ae_loss}
\ell_{2} = \sum_{i \in \mathcal{D}} \parallel x_{i}-\hat{x}_{i}\parallel_2
\end{equation}
The resulting latent space can, for example, be used for unsupervised clustering.

\textbf{Denoising Autoencoders} introduced by \cite{vincent2010stacked} have a modified training procedure. Here, artificial random noise is applied to the input images $x \in \mathcal{R}^{\mathcal{D}}$ while the reconstruction target stays clean. The trained model can be used to reconstruct denoised test images. But how is the latent representation affected? 

\textbf{Hypothesis 1:} \textit{The Denoising AE produces latent representations which are invariant to noise because it facilitates the reconstruction of de-noised images.}

We will demonstrate that this training strategy actually enforces invariance not only against noise but against a variety of different input augmentations. Finally, it allows us to bridge the domain gap between simulated and real data.

\subsection{Augmented Autoencoder}
\label{sec:aae}

\begin{figure*}[t]
	\centering
	\captionsetup{width=\textwidth}
	\includegraphics[width=\textwidth]{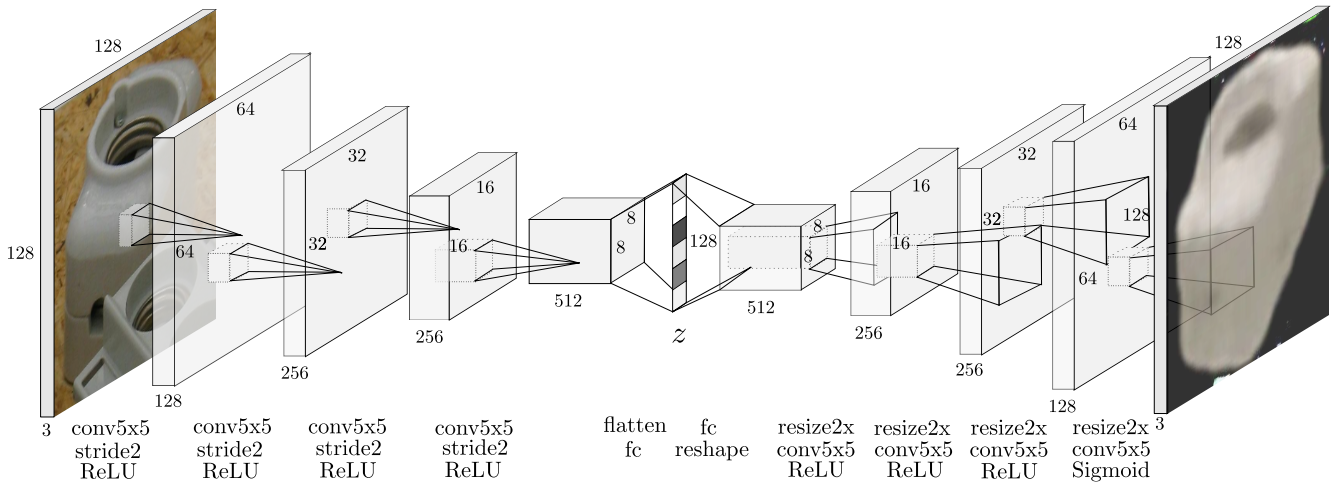}
	\caption{Autoencoder CNN architecture with occluded test input, "resize2x" depicts nearest-neighbor upsampling}
	\label{fig:ae_arch}
\end{figure*}

The motivation behind the \gls{AAE} is to control what the latent representation encodes and which properties are ignored. We apply random augmentations $f_{augm}(.)$ to the input images $x \in \mathcal{R}^{\mathcal{D}}$ against which the encoding should become invariant. The reconstruction target remains eq. \eqref{eq:ae_loss} but eq. \eqref{eq:yhat} becomes
\begin{equation}
\hat{x} = (\Psi\circ\Phi\circ f_{augm})(x) = (\Psi\circ\Phi)(x') = \Psi(z')
\end{equation}
To make evident that \textbf{Hypothesis 1} holds for geometric transformations, we learn latent representations of binary images depicting a 2D square at different scales, in-plane translations and rotations. Our goal is to encode only the in-plane rotations $r \in [0,2 \pi]$ in a two dimensional latent space $z \in \mathcal{R}^{2}$ independent of scale or translation. Fig. \ref{fig:toy} depicts the results after training a \gls{CNN}-based \gls{AE} architecture similar to the model in Fig. \ref{fig:ae_arch}. It can be observed that the \glspl{AE} trained on reconstructing squares at fixed scale and translation (1) or random scale and translation (2) do not clearly encode rotation alone, but are also sensitive to other latent factors. Instead, the encoding of the \gls{AAE} (3) becomes invariant to translation and scale such that all squares with coinciding orientation are mapped to the same code. Furthermore, the latent representation is much smoother and the latent dimensions imitate a shifted sine and cosine function with frequency $f=\frac{4}{2 \pi}$ respectively. The reason is that the square has two perpendicular axes of symmetry, i.e. after rotating $\frac{\pi}{2}$ the square appears the same. This property of representing the orientation based on the appearance of an object rather than on a fixed parametrization is valuable to avoid ambiguities due to symmetries when teaching 3D object orientations.

\subsection{Learning 3D Orientation from Synthetic Object Views}
Our toy problem showed that we can explicitly learn representations of object in-plane rotations using a geometric augmentation technique. Applying the same geometric input augmentations we can encode the whole SO(3) space of views from a 3D object model (CAD or 3D reconstruction) while being robust against inaccurate object detections. However, the encoder would still be unable to relate image crops from real RGB sensors because 
(1) the 3D model and the real object differ, (2) simulated and real lighting conditions differ, (3) the network can't distinguish the object from background clutter and foreground occlusions. 
Instead of trying to imitate every detail of specific real sensor recordings in simulation we propose a Domain Randomization (DR) technique within the \gls{AAE} framework to make the encodings invariant to insignificant environment and sensor variations. The goal is that the trained encoder treats the differences to real camera images as just another irrelevant variation. Therefore, while keeping reconstruction targets clean, we randomly apply additional augmentations to the input training views: (1) rendering with random light positions and randomized diffuse and specular reflection (simple Phong model \citep{phong1975illumination} in OpenGL), (2) inserting random background images from the Pascal VOC dataset \citep{pascalvoc2012}, (3) varying image contrast, brightness, Gaussian blur and color distortions, (4) applying occlusions using random object masks or black squares. Fig. \ref{fig:training_process} depicts an exemplary training process for synthetic views of object 5 from T-LESS \citep{hodan2017tless}.

\begin{table}[t]
	\scriptsize
	\centering
	\captionsetup{width=0.9\columnwidth}
	\caption{Augmentation Parameters of \gls{AAE}; Scale and translation is in relation to image shape and occlusion is in proportion of the object mask}
	\begin{tabular}{cc|cc}
		\toprule
		& 50\% chance &\multicolumn{2}{c}{light (random position) } \\
		& (30\% per channel) & \multicolumn{2}{c}{\& geometric}\\
		\midrule
		add & $\mathcal{U}(-0.1,0.1)$ & ambient &$0.4$ \\
		contrast & $\mathcal{U}(0.4,2.3)$ &diffuse &$\mathcal{U}(0.7,0.9)$\\
		multiply & $\mathcal{U}(0.6,1.4)$ & specular&$\mathcal{U}(0.2,0.4)$ \\
		invert &  & scale &$\mathcal{U}(0.8,1.2)$\\
		gaussian blur & $\sigma \sim \mathcal{U}(0.0,1.2)$ & translation & $\mathcal{U}(-0.15,0.15)$\\
		&& occlusion & $\in [0,0.25]$ 
	\end{tabular}
	\label{tab:aug_strong_col}
\end{table}

\begin{figure}[t]
	\centering
	
	\centering
	\captionsetup{width=0.95\columnwidth}
	\includegraphics[width=0.82\columnwidth]{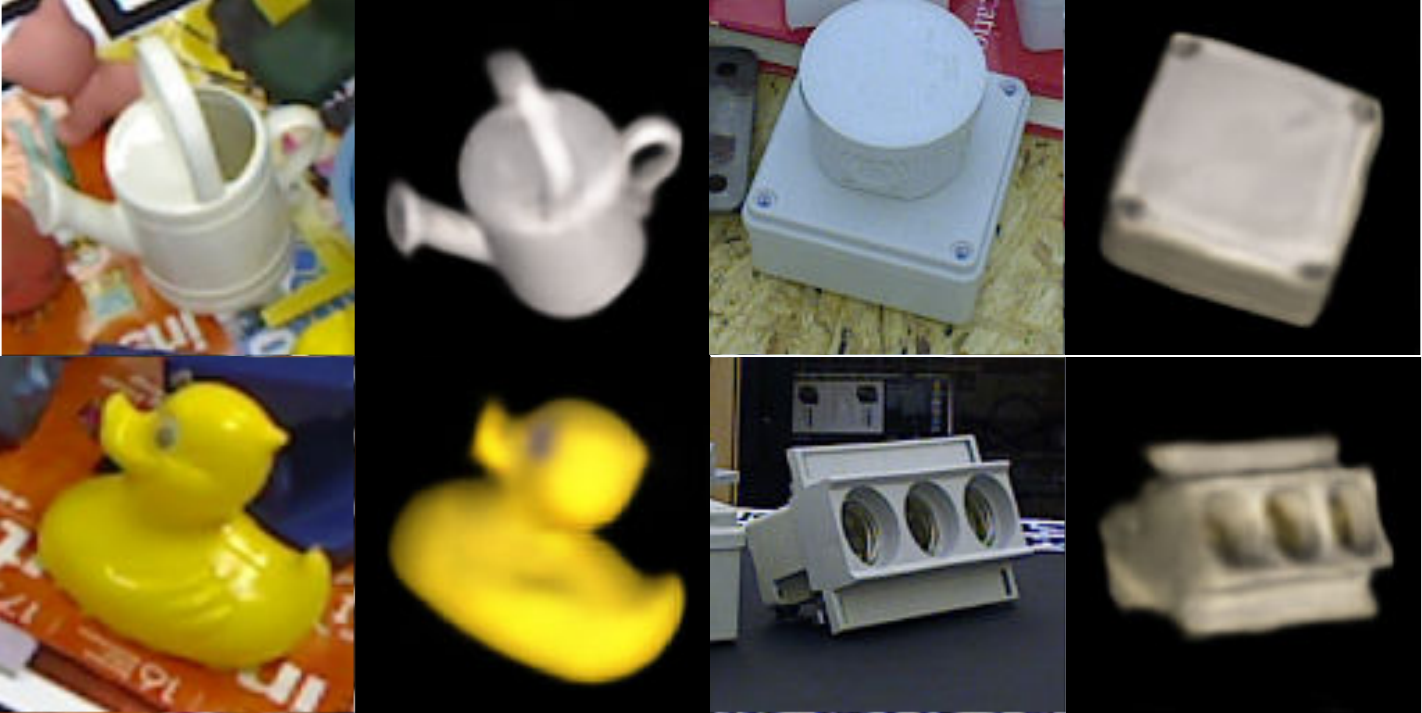}
	\caption{\gls{AAE} decoder reconstruction of LineMOD (left) and T-LESS (right) scene crops}
	\label{fig:reconst_ex}
	
\end{figure}
\subsection{Network Architecture and Training Details}

The convolutional Autoencoder architecture that is used in our experiments is depicted in Fig. \ref{fig:ae_arch}. 
We use a bootstrapped pixel-wise L2 loss, first introduced by \cite{wu2016bridging}. Only the pixels with the largest reconstruction errors contribute to the loss. Thereby, finer details are reconstructed and the training does not converge to local minima like reconstructing black images for all views. In our experiments, we choose a bootstrap factor of $k=4$ per image, meaning that $\frac{1}{4}$ of all pixels contribute to the loss.
Using OpenGL, we render 20000 views of each object uniformly at random 3D orientations and constant distance along the camera axis (700mm). The resulting images are quadratically cropped using the longer side of the bounding box and resized (nearest neighbor) to $128 \times 128 \times 3$ as shown in Fig. \ref{fig:training_process}. All geometric and color input augmentations besides the rendering with random lighting are applied online during training at uniform random strength, parameters are found in Tab. \ref{tab:aug_strong_col}.
We use the Adam \citep{kingma2014adam} optimizer with a learning rate of $2\times 10^{-4}$, Xavier initialization \citep{glorot2010understanding}, a \mbox{batch size = 64} and 40000 iterations which takes $\sim 4$ hours on a single Nvidia Geforce GTX 1080.
\begin{figure*}[t]
	\centering
	\captionsetup{width=0.93\textwidth}
	\includegraphics[width=0.9\textwidth]{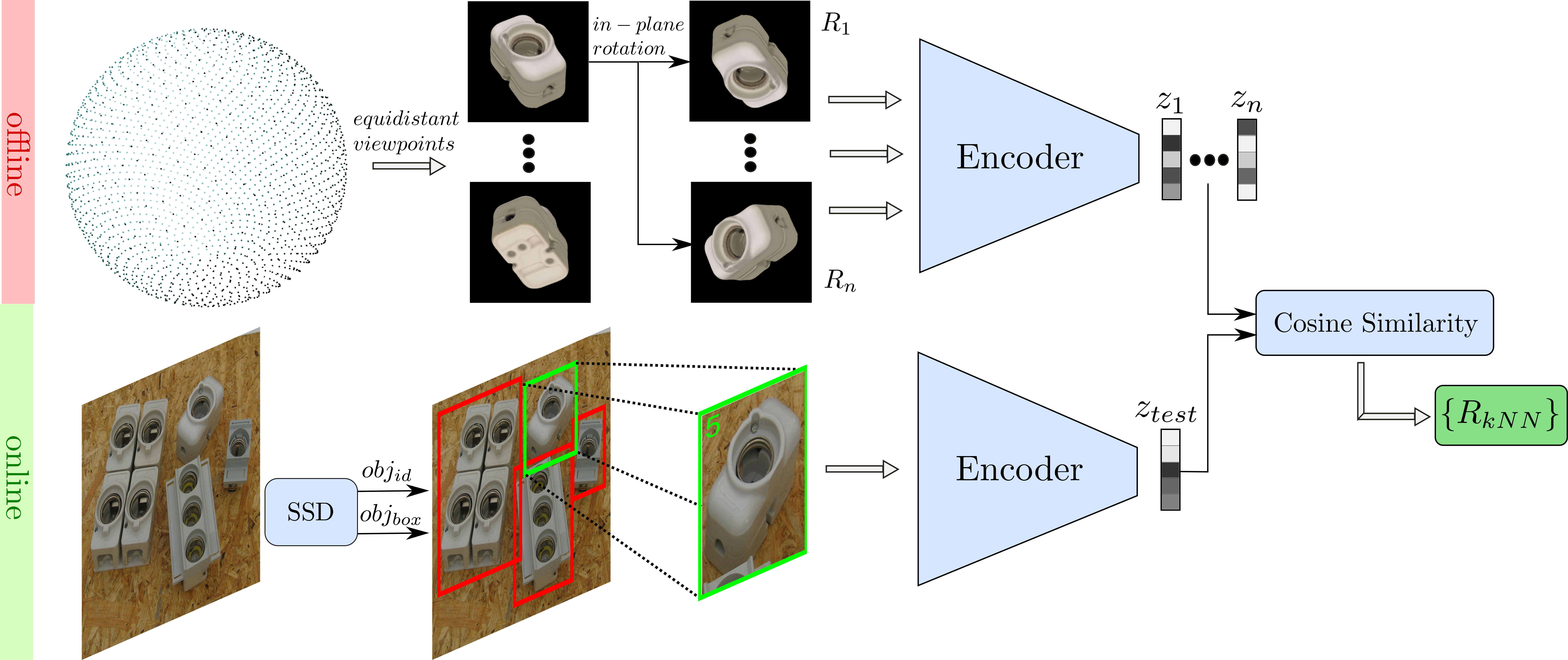}
	\caption{Top: creating a codebook from the encodings of discrete synthetic object views; bottom: object detection and 3D orientation estimation using the nearest neighbor(s) with highest cosine similarity from the codebook}
	\label{fig:test_pipeline}
\end{figure*}
\subsection{Codebook Creation and Test Procedure}

After training, the \gls{AAE} is able to extract a 3D object from real scene crops of many different camera sensors (Fig. \ref{fig:reconst_ex}). The clarity and orientation of the decoder reconstruction is an indicator of the encoding quality. To determine 3D object orientations from test scene crops we create a codebook (Fig. \ref{fig:test_pipeline} (top)):
\begin{itemize}
	\item[1)]  Render clean, synthetic object views at nearly equidistant viewpoints from a full view-sphere (based on a refined icosahedron \citep{hinterstoisser2008simultaneous})
	\item[2)] Rotate each view in-plane at fixed intervals to cover the whole SO(3)
	\item[3)] Create a codebook by generating latent codes $z \in \mathcal{R}^{128}$ for all resulting images and assigning their corresponding rotation $R_{cam2obj} \in \mathcal{R}^{3x3}$
\end{itemize}

\begin{figure}[t]
	\centering
		\captionsetup{width=0.99\columnwidth}
		\includegraphics[width=0.7\columnwidth]{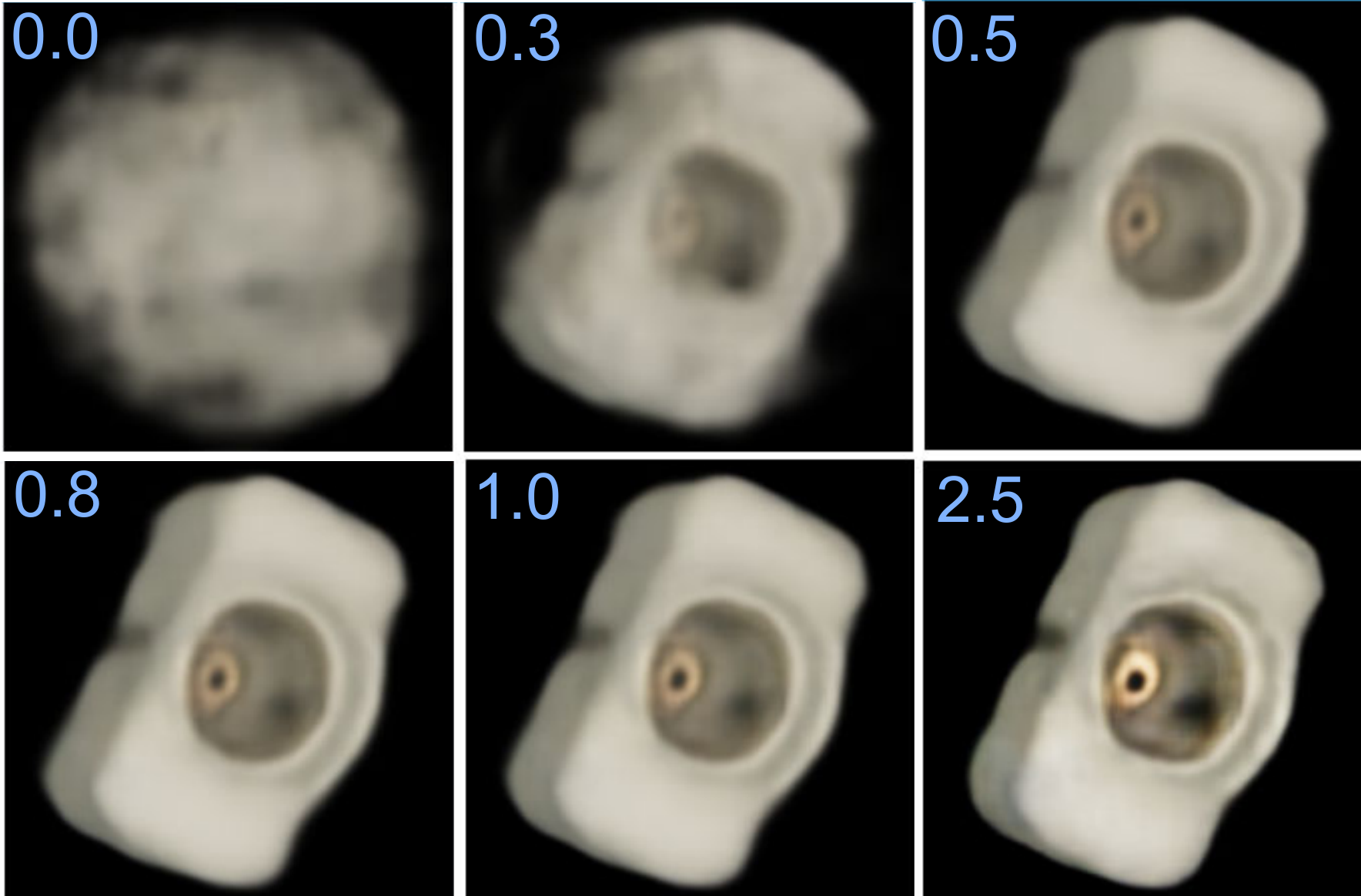}
		\caption{\gls{AAE} decoder reconstruction of a test code $z_{test} \in \mathcal{R}^{128}$ scaled by a factor $s\in[0,2.5]$}
		\label{fig:scale}
\end{figure}

At test time, the considered object(s) are first detected in an RGB scene. The image is quadratically cropped using the longer side of the bounding box multiplied with a padding factor of 1.2 and resized to match the encoder input size. The padding accounts for imprecise bounding boxes. After encoding we compute the cosine similarity between the test code $z_{test} \in \mathcal{R}^{128}$ and all codes $z_{i} \in \mathcal{R}^{128}$ from the codebook:
\begin{equation}
cos_i = \frac{\pmb z_i \;\pmb z_{test}}{\lVert \pmb z_i \rVert \lVert \pmb z_{test} \rVert}
\end{equation}
The highest similarities are determined in a \gls{kNN} search and the corresponding rotation matrices $ \{R_{kNN}\}$ from the codebook are returned as estimates of the 3D object orientation. For the quantitative evaluation we use $k=1$, however the next neighbors can yield valuable information on ambiguous views and could for example be used in particle filter based tracking. We use cosine similarity because (1) it can be very efficiently computed on a single GPU even for large codebooks. In our experiments we have 2562 equidistant viewpoints $\times$ 36 in-plane rotation = 92232 total entries. (2) We observed that, presumably due to the circular nature of rotations, scaling a latent test code does not change the object orientation of the decoder reconstruction (Fig. \ref{fig:scale}).

\begin{table}[t]
	\scriptsize
	\centering
	\captionsetup{width=0.9\columnwidth}
	\caption{Augmentation Parameters for Object Detectors, top five are applied in random order; bottom part describes phong lighting from random light positions}
	\begin{tabular}{cc|cc}
		\toprule
		&chance & SIXD train & Rendered \\
		&(per ch.)& & 3D models \\
		\midrule
		add & 0.5 (0.15) & $\mathcal{U}(-0.08,0.08)$  &$\mathcal{U}(-0.1,0.1)$ \\
		contrast norm. &0.5 (0.15)& $\mathcal{U}(0.5,2.2)$  &$\mathcal{U}(0.5,2.2)$\\
		multiply & 0.5 (0.25) & $\mathcal{U}(0.6,1.4)$ &$\mathcal{U}(0.5,1.5)$ \\
		gaussian blur & 0.2 & $\sigma \sim \mathcal{U}(0.5,1.0)$  & $\sigma = 0.4$ \\
		gaussian noise & 0.1 (0.1) & $\sigma = 0.04$ & - \\
		\midrule
		ambient & 1.0 & -&$0.4$ \\
		diffuse &1.0&-&$\mathcal{U}(0.7,0.9)$ \\
		specular &1.0&-&$\mathcal{U}(0.2,0.4)$ 
	\end{tabular}
	\label{tab:aug_det}
\end{table}
\subsection{Extending to 6D Object Detection}
\subsubsection{Training the 2D Object Detector.} We finetune the 2D Object Detectors using the object views on black background  which are provided in the training datasets of LineMOD and T-LESS. In LineMOD we additionally render domain randomized views of the provided 3D models and freeze the backbone like in \cite{hinterstoisser2017pre}. Multiple object views are sequentially copied into an empty scene at random translation, scale and in-plane rotation. Bounding box annotations are adapted accordingly. If an object view is more than 40\% occluded, we re-sample it. Then, as for the \gls{AAE}, the black background is replaced with Pascal VOC images. The randomization schemes and parameters can be found in Table \ref{tab:aug_det}. In T-LESS we train SSD \citep{liu2016ssd} with VGG16 backbone and RetinaNet \citep{lin2018focal} with ResNet50 backbone which is slower but more accurate, on LineMOD we only train RetinaNet. For T-LESS we generate 60000 training samples from the provided training dataset and for LineMOD we generate 60000 samples from the training dataset plus 60000 samples from 3D model renderings with randomized lighting conditions (see Table \ref{tab:aug_det}). The RetinaNet achieves 0.73mAP@0.5IoU on T-LESS and 0.62mAP@0.5IoU on LineMOD. On Occluded LineMOD, the detectors trained on the simplistic renderings failed to achieve good detection performance. However, recent work of  \cite{Hodan2019PhotorealisticIS} quantitatively investigated the training of 2D detectors on synthetic data and they reached decent detection performance on Occluded LineMOD by fine-tuning FasterRCNN on photo-realistic synthetic images showing the feasibility of a purely synthetic pipeline.

\subsubsection{Projective Distance Estimation}
\label{sec:proj}
We estimate the full 3D translation $t_{real}$ from camera to object center, similar to \cite{kehl2017ssd}. Therefore, we save the 2D bounding box for each synthetic object view in the codebook and compute its diagonal length $\| bb_{syn,i} \|$. At test time, we compute the ratio between the detected bounding box diagonal $\| bb_{real} \|$ and the corresponding codebook diagonal $\| bb_{syn,{\argmax} (cos_i) }\|$, i.e. at similar orientation. The pinhole camera model yields the distance estimate $\hat{t}_{real,z}$
\begin{equation}
\label{eq:dist}
\hat{t}_{real,z} = t_{syn,z} \times   \frac{\| bb_{syn,{\argmax} (cos_i)} \|}{\| bb_{real} \|} \times \frac{f_{real}}{f_{syn}}
\end{equation} 
with synthetic rendering distance $t_{syn,z}$ and focal lengths $f_{real}$, $f_{syn}$ of the real sensor and synthetic views. It follows that

\begin{align}
\boldsymbol{\Delta \hat{t}} &= \hat{t}_{real,z} \boldsymbol{K_{real}^{-1}} \boldsymbol{bb_{real,c}} - t_{syn,z}\boldsymbol{ K_{syn}^{-1}} \boldsymbol{bb_{syn,c}}
\\[+0.5em]
\boldsymbol{\hat{t}_{real}} &= \boldsymbol{t_{syn}} + \boldsymbol{\Delta \hat{t}}
\end{align}

where $\boldsymbol{\Delta \hat{t}}$ is the estimated vector from the synthetic to the real object center, $\boldsymbol{K_{real}}, \boldsymbol{K_{syn}}$ are the camera matrices, $\boldsymbol{bb_{real,c}},\boldsymbol{bb_{syn,c}}$ are the bounding box centers in homogeneous coordinates and $\boldsymbol{\hat{t}_{real}}, \boldsymbol{t_{syn}} = (0,0,t_{syn,z})$ are the translation vectors from camera to object centers. In contrast to \cite{kehl2017ssd}, we can predict the 3D translation for different test intrinsics.

\subsubsection{Perspective Correction}
While the codebook is created by encoding centered object views, the test image crops typically do not originate from the image center. Naturally, the appearance of the object view changes when translating the object in the image plane at constant object orientation. This causes a noticeable error in the rotation estimate from the codebook towards the image borders. However, this error can be corrected by determining the object rotation that approximately preserves the appearance of the object when translating it to our estimate $\boldsymbol{\hat{t}_{real}}$.
\begin{align}
\label{eq:correct}
\left( \begin{array}{c}\alpha_x\\\alpha_y \end{array} \right) 
 &= 
 \left( \begin{array}{c} -\arctan(\hat{t}_{real,y} / \hat{t}_{real,z})  \\ \arctan(\hat{t}_{real,x} / \sqrt{\hat{t}_{real,z}^2 + \hat{t}^2_{real,y}}) \end{array} \right)\\[+1em]
 \boldsymbol{\hat{R}_{obj2cam}} &= \boldsymbol{ R_y(}\alpha_y\boldsymbol{)R_x(}\alpha_x\boldsymbol{)\hat{R}'_{obj2cam}}
\end{align}
where $\alpha_x, \alpha_y$ describe the angles around the camera axes and $\boldsymbol{ R_y(}\alpha_y\boldsymbol{)R_x(}\alpha_x\boldsymbol{)}$ the corresponding rotation matrices to correct the initial rotation estimate $\boldsymbol{\hat{R}'_{obj2cam}}$ from object to camera.
The perspective corrections give a notable boost in accuracy as reported in Table \ref{tab:persp}. If strong perspective distortions are expected at test time, the training images $x'$ could also be recorded at random distances as opposed to constant distance. However, in the benchmarks, perspective distortions are minimal and consequently random online image-plane scaling of $x'$ is sufficient. 

\subsubsection{ICP Refinement}
\label{sec:icp}
Optionally, the estimate is refined on depth data using a point-to-plane \gls{ICP} approach with adaptive thresholding of correspondences based on \cite{chen1992object, zhang1994iterative} taking an average of $\sim 320ms$. The refinement is first applied in direction of the vector pointing from camera to the object where most of the RGB-based pose estimation errors stem from and then on the full 6D pose.

\subsubsection{Inference Time}
The \gls{SSD} with VGG16 base and 31 classes plus the \gls{AAE} (Fig. \ref{fig:ae_arch}) with a codebook size of $92232 \times 128$ yield the average inference times depicted in Table \ref{tab:infer}. We conclude that the RGB-based pipeline is real-time capable at $\sim$42Hz on a Nvidia GTX 1080. This enables augmented reality and robotic applications and leaves room for tracking algorithms. Multiple encoders (15MB) and corresponding codebooks (45MB each) fit into the GPU memory, making multi-object pose estimation feasible.

\begin{table}[t]
		\scriptsize
		\centering
		\captionsetup{width=0.8\columnwidth}
		\caption{Inference time of the RGB pipeline using SSD on CPUs \textbf{or} GPU}
		\begin{tabular}{ccc}
			\toprule
			& 4 CPUs & GPU\\
			\midrule
			SSD & - & \textbf{$\sim$17ms}\\
			Encoder & $\sim$100ms & \textbf{$\sim$5ms}\\
			Cosine Similarity & 2.5ms & \textbf{1.3ms}\\
			Nearest Neighbor &	\textbf{0.3ms} &	3.2ms\\
			Projective Distance &	\textbf{0.4ms} &	-\\
			\midrule
			&\multicolumn{2}{c}{\textbf{$\sim$24ms}}\\
		\end{tabular}
		\label{tab:infer}
\end{table}
\begin{table}[t]
		\centering
		\scriptsize
		\captionsetup{width=0.8\columnwidth}
		\caption{Single object pose estimation runtime w/o refinement}
		\begin{tabular}{cc}
			\toprule
			Method & fps \\
			\midrule
			\cite{vidal20186d} & $0.2$\\
				\\[-0.6em]
			\cite{brachmann2016uncertainty} & 2 \\
			\\[-0.6em]
			\cite{kehl2016deep} & 2 \\
			\\[-0.6em]
			\cite{rad2017bb8} & 4 \\
			\\[-0.6em]
			\cite{kehl2017ssd} & 12 \\
			\\[-0.6em]
			OURS & 13 (RetinaNet) \\
			\\[-0.6em]
				&  42 (SSD) \\
			\\[-0.6em]
			\cite{tekin2017real} & 50 \\
		\end{tabular}
		\label{tab:runtime_compare}

\end{table}

\begin{table*}[t]
	\centering
	\scriptsize
	\captionsetup{justification=centering,width=.8\textwidth}
	\caption{Ablation study on color augmentations for different test sensors. Object 5 tested on all scenes, T-LESS \cite{hodan2017tless}. Standard deviation of three runs in brackets.}
	\begin{adjustbox}{max width=.8\textwidth}
		\begin{tabular}{ccccccccc}
			\toprule
			\textbf{Train RGB} & \textbf{Test RGB}
			& dyn. light & add & contrast & multiply & invert & \textbf{AUC\textsubscript{vsd}}\\
			\midrule
			3D Reconstruction & Primesense &  \OK  &  &  &  &  & 0.472 ($\pm$ 0.013)\\
			(synthetic)& (real) & \OK  & \OK &  &  &   & 0.611 ($\pm$ 0.030)\\
			&& \OK  & \OK & \OK &  &  & 0.825 ($\pm$ 0.015)\\
			&& \OK  & \OK & \OK &  \OK&   & 0.876 ($\pm$ 0.019)\\
			&& \OK  & \OK & \OK &  \OK&  \OK & \textbf{0.877} ($\pm$ 0.005)\\
			&&    & \OK & \OK &  \OK&  & 0.861 ($\pm$ 0.014)\\   
			\midrule
			Primesense (real) & Primesense (real) & & \OK & \OK &  \OK& & 0.890 ($\pm$ 0.003)\\
			\midrule
			3D Reconstruction&Kinect &  \OK  &  &  &  &  & 0.461 ($\pm$ 0.022)\\
			(synthetic)&(real) &  \OK  & \OK &  &  &   & 0.580 ($\pm$ 0.014)\\
			&&  \OK  & \OK & \OK &  &  & 0.701 ($\pm$ 0.046)\\
			&&  \OK  & \OK & \OK &  \OK&   & 0.855 ($\pm$ 0.016)\\
			&&  \OK  & \OK & \OK &  \OK&  \OK & 0.897 ($\pm$ 0.008)\\
			&&    & \OK & \OK &  \OK&   & \textbf{0.903} ($\pm$ 0.016)\\
			\midrule
			Kinect (real)& Kinect (real) & & \OK & \OK &  \OK& & 0.917 ($\pm$ 0.007)\\
		\end{tabular}
	\end{adjustbox}
	\label{tab:auc_aug}
\end{table*}
\begin{figure*}[t]
	
	\centering
	
	\subfloat[Effect of latent space size, standard deviation in red \label{fig:latent}]{{\includegraphics[width=0.3\textwidth]{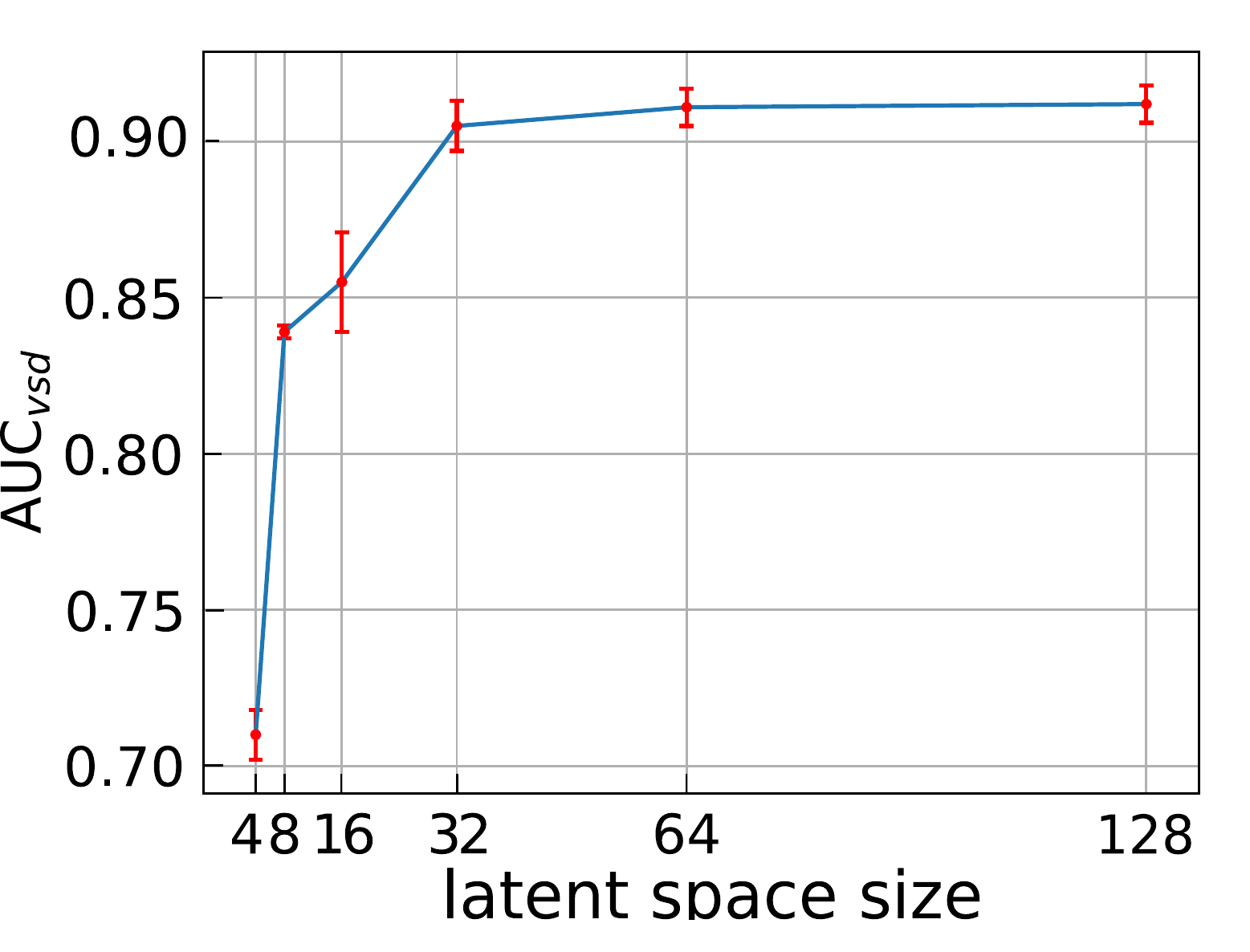}}} \qquad
	\subfloat[Training on CAD model (bottom) vs. textured 3D reconstruction (top) \label{fig:cad_vs_reconst}]{{\includegraphics[width=0.44\textwidth]{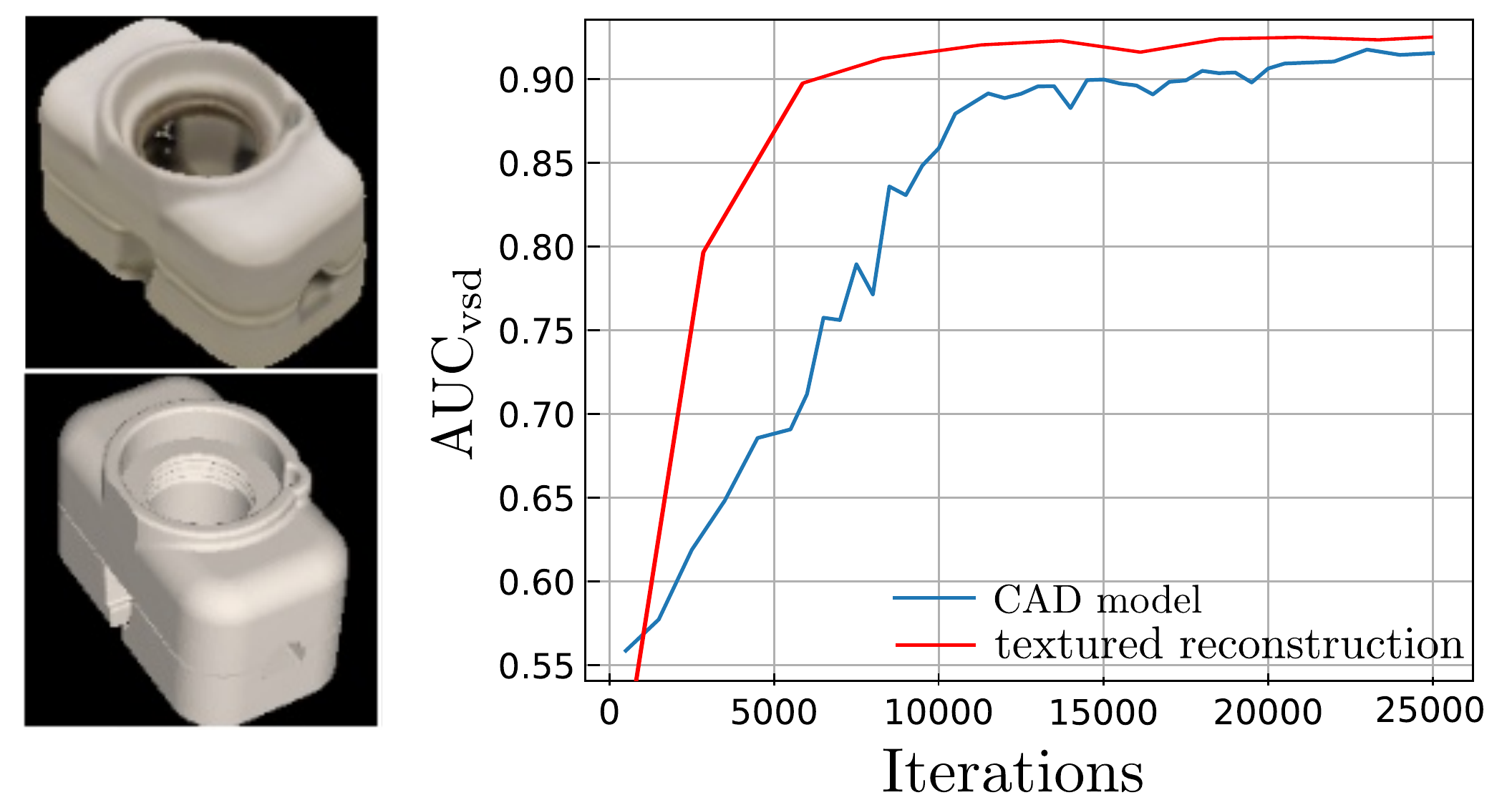}}}
	\captionsetup{}
	\caption{Testing object 5 on all 504 Kinect RGB views of scene 2 in T-LESS}
	
\end{figure*}

\section{Evaluation}

We evaluate the \gls{AAE} and the whole 6D detection pipeline on the T-LESS \citep{hodan2017tless} and LineMOD \citep{hinterstoisser2011multimodal} datasets.
\subsection{Test Conditions}
\label{sec:cond}
Few RGB-based pose estimation approaches (e.g. \cite{kehl2017ssd,ulrich2009cad}) only rely on 3D model information. Most methods like \cite{wohlhart2015learning,balntas2017pose,brachmann2016uncertainty} make use of real pose annotated data and often even train and test on the same scenes (e.g. at slightly different viewpoints, as in the official LineMOD benchmark). It is common practice to ignore in-plane rotations or to only consider object poses that appear in the dataset \citep{rad2017bb8,wohlhart2015learning} which also limits applicability. Symmetric object views are often individually treated \citep{rad2017bb8,balntas2017pose} or ignored \citep{wohlhart2015learning}. 
The SIXD challenge \citep{sixd} is an attempt to make fair comparisons between 6D localization algorithms by prohibiting the use of test scene pixels. We follow these strict evaluation guidelines, but treat the harder problem of 6D detection where it is unknown which of the considered objects are present in the scene. This is especially difficult in the T-LESS dataset since objects are very similar. 
We train the AAEs on the reconstructed 3D models, except for object 19-23 where we train on the CAD models because the pins are missing in the reconstructed plugs. 

We noticed, that the geometry of some 3D reconstruction in T-LESS is slightly inaccurate which badly influences the RGB-based distance estimation (Sec. \ref{sec:proj}) since the synthetic bounding box diagonals are wrong. Therefore, in a second training run we only train on the 30 CAD models.

{\setlength{\tabcolsep}{0.2em}
	\begin{table*}[t]
	\scriptsize
	\centering
	\captionsetup{justification=centering}
	\caption{T-LESS: Object recall for $err_{vsd}<0.3$ on all Primesense test scenes (SIXD/BOP benchmark from \cite{hodan2018bop}). $RGB^\dag$ depicts training with 3D reconstructions, except objects 19-23 $\longrightarrow$ CAD models; $RGB^\ddag$ depicts training on untextured CAD models only }
	\begin{adjustbox}{max width=\textwidth}
		\begin{tabular}{r|ccc|ccccc|cc}
			\toprule
			& \multicolumn{3}{c|}{\textbf{AAE}} &\textbf{AAE}& &&&  & \multicolumn{2}{|c}{\textbf{AAE}}\\
			
			& SSD& \multicolumn{2}{|c|}{RetinaNet}& RetinaNet & \citeauthor{brachmann2016uncertainty}&  \citeauthor{kehl2016deep} & \citeauthor{vidal20186d} & \citeauthor{drost2010model} & \multicolumn{2}{|c}{w/ GT 2D BBs}\\
			
			Data & $RGB^\dag$ & $RGB^\dag$ &$RGB^\ddag$ & $RGB^\dag$+Depth(ICP)  & RGB-D &RGB-D +ICP  & Depth +ICP & Depth +edge & $RGB^\dag$ & +Depth(ICP)\\     
			\midrule                                                  
			1 &       5.65 &   		9.48 &   12.67 &   67.95  & 8&  7 &  43 &   53 & 12.67 &     85.98 \\
			2 &       5.46 &   		13.24 &  16.01 &   70.62  & 10&     10 &  47 &   44 & 11.47 &     86.27 \\
			3 &       7.05 &   		12.78 &  22.84 &   78.39  & 21&     18 &  69 &   61 & 13.32 &     90.80 \\
			4 &       4.61 &   		6.66 &   6.70 &    57.00  & 4&  24 &  63 &   67 & 12.88 &     84.20 \\
			5 &      36.45 &   		36.19 &  38.93 &   77.18  & 46&     23 &  69 &   71 & 67.37 &     90.14 \\
			6 &      23.15 &   		20.64 &  28.26 &   72.75  & 19&     10 &  67 &   73 & 54.21 &     90.58 \\
			7 &      15.97 &   		17.41 &  26.56 &   83.39  & 52&     0 &  77 &   75 & 38.10 &     86.94 \\
			8 &       10.86 &  		21.72 &  18.01 &   78.08  & 22&     2 &  79 &  89 & 24.83 &     91.79 \\
			9 &      19.59 &   		39.98 &  33.36 &   88.64  & 12&     11 &  90 &   92 & 49.06 &     91.09 \\
			10 &       10.47 & 		13.37 &  33.15 &   84.47  & 7&      17 &  68 &  72 & 15.67 &     84.67 \\
			11 &       4.35 &  		7.78 &   17.94 &   56.01  & 3&  5 &  69 &   64 & 16.64 &     77.01 \\
			12 &       7.80 &  		9.54 &   18.38 &   63.23  & 3&  1 &  82 &   81 & 33.57 &     79.32 \\
			13 &       3.30 &  		4.56 &   16.20 &   43.55  & 0&  0 &  56 &   53 & 15.29 &     64.38 \\
			14 &       2.85 &  		5.36 &   10.58 &   25.58  & 0&  9 &  47 &   46 & 50.14 &     71.37 \\
			15 &       7.90 &  		27.11 &  40.50 &   69.81  & 0&  12 &  52 &   55 & 52.01 &     73.90 \\
			16 &      13.06 &  		22.04 &  35.67 &   84.55  & 5&  56 &  81 &   85 & 36.71 &     87.58 \\
			17 &      41.70 &  		66.33 &  50.47 &   74.29  & 3&  52 &  83 &   88 & 81.44 &     78.88 \\
			18 &      47.17 &  		14.91 &  33.63 &   83.12  & 54&     22 &  80 &   78 & 55.48 &     85.64 \\
			19 &       15.95 & 		23.03 &  23.03 &   58.13  & 38&     35 &  55 &  55 & 53.07 &     82.71 \\
			20 &       2.17 &  		5.35 &   5.35 &    26.73  & 1&  5 &  47 &   47 & 38.97 &     70.87 \\
			21 &       19.77 & 		19.82 &  19.82 &   53.48  & 39&     26 &  63 &  55 & 53.45 &     86.83 \\
			22 &       11.01 & 	    20.25 &  20.25 &   60.49  & 19&     27 & 70 &  56 & 49.95 &     84.20 \\
			23 &       7.98 &  		19.15 &  19.15 &   62.69  & 61&     71 &  85 &  84 & 36.74 &     76.40 \\
			24 &       4.74 &  		4.54 &   27.94 &   62.99  & 1&  36 &  70 &   59 & 11.75 &     84.38 \\
			25 &      21.91 &  		19.07 &  51.01 &   73.33  & 16&     28 &  48 &   47 & 37.73 &     87.53 \\
			26 &       10.04 & 		12.92 &  33.00 &   67.00  & 27&     51 &  55 &  69 & 29.82 &     90.26 \\
			27 &       7.42 &  		22.37 &  33.61 &   82.16  & 17&     34 &  60 &   61 & 23.30 &     84.43 \\
			28 &      21.78 &  		24.00 &  30.88 &   83.51  & 13&     54 &  69 &   80 & 43.97 &     89.84 \\
			29 &      15.33 &  		27.66 &  35.57 &   74.45  & 6&  86 &  65 &   84 & 57.82 &     88.58 \\
			30 &      34.63 &  		30.53 &  44.33 &   93.65  & 5&  69 &  84 &   89 & 72.81 &     95.01 \\
			\midrule
			Mean &14.67 &19.26 & \textbf{26.79} &\textbf{68.57}& 17.84 &24.60& 66.51 &  67.50 &38.34 &\textbf{84.05}\\
			\midrule
			Time(s) & \textbf{0.024} &  0.077 &  0.077 &\textbf{0.4}& 13.5 & 1.8& 4.7 &  21.5 & \textbf{0.006} &0.33
		\end{tabular}
	\end{adjustbox}
	
	\label{tab:tless}
\end{table*}

	{\setlength{\tabcolsep}{0.4em}
	\begin{table}
		\centering
		\small
		\captionsetup{width=\textwidth}
		\caption{Effect of Perspective Corrections on T-LESS}
		\begin{tabular}{c|c}
			\toprule
			Method & $RGB^\dag$  \\
			\midrule
			w/o correction & 18.35 \\
			w/ correction & \textbf{19.26 (+0.91)} 
		\end{tabular}
		\label{tab:persp}
		
	\end{table}

\subsection{Metrics}

\cite{hodan2016evaluation}  introduced the \gls{VSD}, an ambiguity-invariant pose error function that is determined by the distance between the estimated and ground truth visible object depth surfaces. As in the SIXD challenge, we report the recall of correct 6D object poses at $err_{vsd} < 0.3$ with tolerance $\tau = 20mm$ and $>10\%$ object visibility. Although the Average Distance of Model Points (ADD) metric introduced by \cite{hinterstoisser2012model} cannot handle pose ambiguities, we also present it for the LineMOD dataset following the official protocol in \cite{hinterstoisser2012model}. For objects with symmetric views (eggbox, glue), \cite{hinterstoisser2012model} adapts the metric by calculating the average distance to the \textit{closest} model point. \cite{Manhardt_2018_ECCV} has noticed inaccurate intrinsics and sensor registration errors between RGB and D in the LineMOD dataset. Thus, purely synthetic RGB-based approaches, although visually correct, suffer from false pose rejections. The focus of our experiments lies on the T-LESS dataset.

 In our ablation studies we also report the $AUC_{vsd}$, which represents the area under the '$err_{vsd}$ vs. recall' curve: 
\begin{align}
AUC_{vsd} = \int_0^1recall(err_{vsd})\,derr_{vsd}
\end{align}

	\subsection{Ablation Studies}
	To assess the \gls{AAE} alone, in this subsection we only predict the 3D orientation of Object 5 from the T-LESS dataset on Primesense and Kinect RGB scene crops. 
	Table \ref{tab:auc_aug} shows the influence of different input augmentations. 
	It can be seen that the effect of different color augmentations is cumulative. For textureless objects, even the inversion of color channels seems to be beneficial since it prevents overfitting to synthetic color information. Furthermore, training with real object recordings provided in T-LESS with random Pascal VOC background and augmentations yields only slightly better performance than the training with synthetic data.
	Fig. \ref{fig:latent} depicts the effect of different latent space sizes on the 3D pose estimation accuracy. Performance starts to saturate at $dim = 64$.

\begin{table*}[t]
	\scriptsize
	\captionsetup{justification=centering,width=0.8\textwidth}
	\caption{LineMOD: Object recall (ADD \cite{hinterstoisser2012model} metric) of methods that use different amounts of training and test data, results taken from \cite{tekin2017real}}
	\begin{adjustbox}{max width=\textwidth}
		\begin{tabular}{r|cc|ccccccc|ccc}
			
			Test data&\multicolumn{9}{c}{RGB}& \multicolumn{3}{|c}{$\qquad$ +Depth (ICP)$\qquad$}\\
			\toprule
			Train data&\multicolumn{2}{c|}{RGB w/o real pose labels}& \multicolumn{7}{c}{RGB with real pose labels} & \multicolumn{2}{|c}{--}\\
			\toprule
			Object &\citeauthor{kehl2017ssd} & OURS & \multicolumn{2}{c}{\citeauthor{brachmann2016uncertainty}} & \multicolumn{2}{c}{\citeauthor{rad2017bb8}} &\citeauthor{tekin2017real} & \multicolumn{2}{c|}{\citeauthor{xiang2017posecnn}} & OURS & \citeauthor{kehl2017ssd} \\
			\toprule
			&&&& +refine && +refine &&& +DeepIm \\
			\midrule
			Ape & 0.00 &   \textbf{4.18} &      -& 	33.2& 27.9& 40.4& 21.62            & -& \textbf{77.0} & 24.35 &\textbf{65}\\
			Benchvise &  0.18 &\textbf{22.85}&  -& 	64.8& 62.0& 91.8& 81.80   & -& \textbf{97.5} & \textbf{89.13}&80\\
			Cam &0.41 & \textbf{32.91}&         -& 	38.4& 40.1& 55.7& 36.57            & -& \textbf{93.5} & \textbf{82.10}&78\\
			Can & 1.35 &   \textbf{37.03}&      -& 	62.9& 48.1& 64.1& 68.80   & -& \textbf{96.5} & 70.82&\textbf{86}\\
			Cat &  0.51 & \textbf{18.68}&       -& 	42.7& 45.2& 62.6& 41.82            & -& \textbf{82.1} & \textbf{72.18}&70\\
			Driller &  2.58 & \textbf{24.81}&   -& 	61.9& 58.6& 74.4& 63.51   & -& \textbf{95.0} & 44.87&\textbf{73}\\
			Duck & 0.00 & \textbf{5.86}&        -& 	30.2& 32.8& 44.3& 27.23            & -& \textbf{77.7} & 54.63 &\textbf{66}\\
			Eggbox & 8.90 & \textbf{81.00}&     -& 	49.9& 40.0& 57.8& 69.58   & -& \textbf{97.1} & 96.62&\textbf{100}\\
			Glue&0.00  & \textbf{46.17}&        -& 	31.2& 27.0& 41.2& 80.02   & -& \textbf{99.4} & 94.18&\textbf{100}\\
			Holepuncher & 0.30 &\textbf{18.20}& -& 	52.8& 42.4& \textbf{67.2}& 42.63   & -& 52.8 & \textbf{51.25}&49\\
			Iron & 8.86 &\textbf{35.05} &       -& 	80.0& 67.0& 84.7& 74.97   & -& \textbf{98.3} & \textbf{77.86}&\textbf{78}\\
			Lamp & 8.2 & \textbf{61.15} &       -& 	67.0& 39.9& 76.5& 71.11   & -& \textbf{97.5} & \textbf{86.31}&73\\
			Phone & 0.18 &\textbf{36.27} &      -& 	38.1& 35.2& 54.0& 47.74   & -& \textbf{87.7} & \textbf{86.24}&79\\
			\midrule
			Mean &  2.42& \textbf{32.63}  & 32.3   & 50.2 & 43.6& 62.7 & 55.95  & 62.7 & \textbf{88.6} & 71.58 & \textbf{79}
		\end{tabular}
	\end{adjustbox}
	\label{tab:linemod}
\end{table*}

\begin{figure*}
	\captionsetup{width=0.8\textwidth}
	\subfloat[Object 5, one view-dependent symmetry]{
		\begin{minipage}[t]{0.5\columnwidth}
\begin{tikzpicture}

\definecolor{color0}{rgb}{0.12156862745098,0.466666666666667,0.705882352941177}
\scriptsize
\begin{axis}[
xlabel={Rotation err [deg]},
ylabel={views},
xmin=0, xmax=180,
ymin=0, ymax=919.8,
width=\columnwidth,
height=\columnwidth,
tick align=outside,
tick pos=left,
x grid style={lightgray!92.02614379084967!black},
y grid style={lightgray!92.02614379084967!black}
]
\draw[fill=color0,draw opacity=0] (axis cs:2.5,0) rectangle (axis cs:7.5,876);
\draw[fill=color0,draw opacity=0] (axis cs:12.5,0) rectangle (axis cs:17.5,80);
\draw[fill=color0,draw opacity=0] (axis cs:22.5,0) rectangle (axis cs:27.5,18);
\draw[fill=color0,draw opacity=0] (axis cs:32.5,0) rectangle (axis cs:37.5,6);
\draw[fill=color0,draw opacity=0] (axis cs:42.5,0) rectangle (axis cs:47.5,5);
\draw[fill=color0,draw opacity=0] (axis cs:52.5,0) rectangle (axis cs:57.5,12);
\draw[fill=color0,draw opacity=0] (axis cs:62.5,0) rectangle (axis cs:67.5,2);
\draw[fill=color0,draw opacity=0] (axis cs:72.5,0) rectangle (axis cs:77.5,5);
\draw[fill=color0,draw opacity=0] (axis cs:82.5,0) rectangle (axis cs:87.5,17);
\draw[fill=color0,draw opacity=0] (axis cs:92.5,0) rectangle (axis cs:97.5,21);
\draw[fill=color0,draw opacity=0] (axis cs:102.5,0) rectangle (axis cs:107.5,27);
\draw[fill=color0,draw opacity=0] (axis cs:112.5,0) rectangle (axis cs:117.5,15);
\draw[fill=color0,draw opacity=0] (axis cs:122.5,0) rectangle (axis cs:127.5,10);
\draw[fill=color0,draw opacity=0] (axis cs:132.5,0) rectangle (axis cs:137.5,18);
\draw[fill=color0,draw opacity=0] (axis cs:142.5,0) rectangle (axis cs:147.5,34);
\draw[fill=color0,draw opacity=0] (axis cs:152.5,0) rectangle (axis cs:157.5,33);
\draw[fill=color0,draw opacity=0] (axis cs:162.5,0) rectangle (axis cs:167.5,109);
\draw[fill=color0,draw opacity=0] (axis cs:172.5,0) rectangle (axis cs:177.5,494);
\end{axis}

\end{tikzpicture}
\begin{tikzpicture}

\definecolor{color0}{rgb}{0.12156862745098,0.466666666666667,0.705882352941177}
\scriptsize
\begin{axis}[
xlabel={Translation err [mm]},
ylabel={views},
xmin=-3.47222222222222, xmax=103.472222222222,
ymin=0, ymax=271.95,
width=\columnwidth,
height=\columnwidth,
xtick={0,25,50,75,100},
tick align=outside,
tick pos=left,
x grid style={lightgray!92.02614379084967!black},
y grid style={lightgray!92.02614379084967!black}
]
\draw[fill=color0,draw opacity=0] (axis cs:1.38888888888889,0) rectangle (axis cs:4.16666666666667,207);
\draw[fill=color0,draw opacity=0] (axis cs:6.94444444444444,0) rectangle (axis cs:9.72222222222222,210);
\draw[fill=color0,draw opacity=0] (axis cs:12.5,0) rectangle (axis cs:15.2777777777778,240);
\draw[fill=color0,draw opacity=0] (axis cs:18.0555555555556,0) rectangle (axis cs:20.8333333333333,259);
\draw[fill=color0,draw opacity=0] (axis cs:23.6111111111111,0) rectangle (axis cs:26.3888888888889,231);
\draw[fill=color0,draw opacity=0] (axis cs:29.1666666666667,0) rectangle (axis cs:31.9444444444444,133);
\draw[fill=color0,draw opacity=0] (axis cs:34.7222222222222,0) rectangle (axis cs:37.5,85);
\draw[fill=color0,draw opacity=0] (axis cs:40.2777777777778,0) rectangle (axis cs:43.0555555555556,47);
\draw[fill=color0,draw opacity=0] (axis cs:45.8333333333333,0) rectangle (axis cs:48.6111111111111,28);
\draw[fill=color0,draw opacity=0] (axis cs:51.3888888888889,0) rectangle (axis cs:54.1666666666667,16);
\draw[fill=color0,draw opacity=0] (axis cs:56.9444444444444,0) rectangle (axis cs:59.7222222222222,20);
\draw[fill=color0,draw opacity=0] (axis cs:62.5,0) rectangle (axis cs:65.2777777777778,6);
\draw[fill=color0,draw opacity=0] (axis cs:68.0555555555555,0) rectangle (axis cs:70.8333333333333,11);
\draw[fill=color0,draw opacity=0] (axis cs:73.6111111111111,0) rectangle (axis cs:76.3888888888889,7);
\draw[fill=color0,draw opacity=0] (axis cs:79.1666666666667,0) rectangle (axis cs:81.9444444444444,7);
\draw[fill=color0,draw opacity=0] (axis cs:84.7222222222222,0) rectangle (axis cs:87.5,8);
\draw[fill=color0,draw opacity=0] (axis cs:90.2777777777778,0) rectangle (axis cs:93.0555555555555,6);
\draw[fill=color0,draw opacity=0] (axis cs:95.8333333333333,0) rectangle (axis cs:98.6111111111111,14);
\end{axis}

\end{tikzpicture}
		\end{minipage}%
		\begin{minipage}[t]{0.5\columnwidth}
\begin{tikzpicture}

\definecolor{color0}{rgb}{0.12156862745098,0.466666666666667,0.705882352941177}
\scriptsize
\begin{axis}[
xlabel={Rotation err [deg]},
xmin=0, xmax=180,
ymin=0, ymax=915.6,
width=\columnwidth,
height=\columnwidth,
tick align=outside,
tick pos=left,
x grid style={lightgray!92.02614379084967!black},
y grid style={lightgray!92.02614379084967!black}
]
\draw[fill=color0,draw opacity=0] (axis cs:2.5,0) rectangle (axis cs:7.5,872);
\draw[fill=color0,draw opacity=0] (axis cs:12.5,0) rectangle (axis cs:17.5,89);
\draw[fill=color0,draw opacity=0] (axis cs:22.5,0) rectangle (axis cs:27.5,13);
\draw[fill=color0,draw opacity=0] (axis cs:32.5,0) rectangle (axis cs:37.5,5);
\draw[fill=color0,draw opacity=0] (axis cs:42.5,0) rectangle (axis cs:47.5,7);
\draw[fill=color0,draw opacity=0] (axis cs:52.5,0) rectangle (axis cs:57.5,11);
\draw[fill=color0,draw opacity=0] (axis cs:62.5,0) rectangle (axis cs:67.5,2);
\draw[fill=color0,draw opacity=0] (axis cs:72.5,0) rectangle (axis cs:77.5,3);
\draw[fill=color0,draw opacity=0] (axis cs:82.5,0) rectangle (axis cs:87.5,15);
\draw[fill=color0,draw opacity=0] (axis cs:92.5,0) rectangle (axis cs:97.5,23);
\draw[fill=color0,draw opacity=0] (axis cs:102.5,0) rectangle (axis cs:107.5,29);
\draw[fill=color0,draw opacity=0] (axis cs:112.5,0) rectangle (axis cs:117.5,16);
\draw[fill=color0,draw opacity=0] (axis cs:122.5,0) rectangle (axis cs:127.5,9);
\draw[fill=color0,draw opacity=0] (axis cs:132.5,0) rectangle (axis cs:137.5,18);
\draw[fill=color0,draw opacity=0] (axis cs:142.5,0) rectangle (axis cs:147.5,32);
\draw[fill=color0,draw opacity=0] (axis cs:152.5,0) rectangle (axis cs:157.5,39);
\draw[fill=color0,draw opacity=0] (axis cs:162.5,0) rectangle (axis cs:167.5,83);
\draw[fill=color0,draw opacity=0] (axis cs:172.5,0) rectangle (axis cs:177.5,515);
\end{axis}

\end{tikzpicture}
\begin{tikzpicture}

\definecolor{color0}{rgb}{0.12156862745098,0.466666666666667,0.705882352941177}
\scriptsize
\begin{axis}[
xlabel={Translation err [mm]},
xmin=-3.47222222222222, xmax=103.472222222222,
ymin=0, ymax=1417.5,
width=\columnwidth,
height=\columnwidth,
xtick={0,25,50,75,100},
ytick={0,300,600,900},
tick align=outside,
tick pos=left,
x grid style={lightgray!92.02614379084967!black},
y grid style={lightgray!92.02614379084967!black}
]
\draw[fill=color0,draw opacity=0] (axis cs:1.38888888888889,0) rectangle (axis cs:4.16666666666667,1350);
\draw[fill=color0,draw opacity=0] (axis cs:6.94444444444444,0) rectangle (axis cs:9.72222222222222,114);
\draw[fill=color0,draw opacity=0] (axis cs:12.5,0) rectangle (axis cs:15.2777777777778,20);
\draw[fill=color0,draw opacity=0] (axis cs:18.0555555555556,0) rectangle (axis cs:20.8333333333333,6);
\draw[fill=color0,draw opacity=0] (axis cs:23.6111111111111,0) rectangle (axis cs:26.3888888888889,2);
\draw[fill=color0,draw opacity=0] (axis cs:29.1666666666667,0) rectangle (axis cs:31.9444444444444,0);
\draw[fill=color0,draw opacity=0] (axis cs:34.7222222222222,0) rectangle (axis cs:37.5,1);
\draw[fill=color0,draw opacity=0] (axis cs:40.2777777777778,0) rectangle (axis cs:43.0555555555556,0);
\draw[fill=color0,draw opacity=0] (axis cs:45.8333333333333,0) rectangle (axis cs:48.6111111111111,2);
\draw[fill=color0,draw opacity=0] (axis cs:51.3888888888889,0) rectangle (axis cs:54.1666666666667,4);
\draw[fill=color0,draw opacity=0] (axis cs:56.9444444444444,0) rectangle (axis cs:59.7222222222222,5);
\draw[fill=color0,draw opacity=0] (axis cs:62.5,0) rectangle (axis cs:65.2777777777778,1);
\draw[fill=color0,draw opacity=0] (axis cs:68.0555555555555,0) rectangle (axis cs:70.8333333333333,1);
\draw[fill=color0,draw opacity=0] (axis cs:73.6111111111111,0) rectangle (axis cs:76.3888888888889,1);
\draw[fill=color0,draw opacity=0] (axis cs:79.1666666666667,0) rectangle (axis cs:81.9444444444444,4);
\draw[fill=color0,draw opacity=0] (axis cs:84.7222222222222,0) rectangle (axis cs:87.5,10);
\draw[fill=color0,draw opacity=0] (axis cs:90.2777777777778,0) rectangle (axis cs:93.0555555555555,5);
\draw[fill=color0,draw opacity=0] (axis cs:95.8333333333333,0) rectangle (axis cs:98.6111111111111,14);
\end{axis}

\end{tikzpicture}
	\end{minipage}}%
	\subfloat[Object 28, two view-dependent symmetries]{
		\begin{minipage}[t]{0.5\columnwidth}
\begin{tikzpicture}

\definecolor{color0}{rgb}{0.12156862745098,0.466666666666667,0.705882352941177}
\scriptsize
\begin{axis}[
xlabel={Rotation err [deg]},
ylabel={views},
xmin=0, xmax=180,
ymin=0, ymax=714,
width=\columnwidth,
height=\columnwidth,
tick align=outside,
tick pos=left,
x grid style={lightgray!92.02614379084967!black},
y grid style={lightgray!92.02614379084967!black}
]
\draw[fill=color0,draw opacity=0] (axis cs:2.36842105263158,0) rectangle (axis cs:7.10526315789474,323);
\draw[fill=color0,draw opacity=0] (axis cs:11.8421052631579,0) rectangle (axis cs:16.5789473684211,63);
\draw[fill=color0,draw opacity=0] (axis cs:21.3157894736842,0) rectangle (axis cs:26.0526315789474,48);
\draw[fill=color0,draw opacity=0] (axis cs:30.7894736842105,0) rectangle (axis cs:35.5263157894737,8);
\draw[fill=color0,draw opacity=0] (axis cs:40.2631578947368,0) rectangle (axis cs:45,7);
\draw[fill=color0,draw opacity=0] (axis cs:49.7368421052632,0) rectangle (axis cs:54.4736842105263,11);
\draw[fill=color0,draw opacity=0] (axis cs:59.2105263157895,0) rectangle (axis cs:63.9473684210526,18);
\draw[fill=color0,draw opacity=0] (axis cs:68.6842105263158,0) rectangle (axis cs:73.421052631579,60);
\draw[fill=color0,draw opacity=0] (axis cs:78.1578947368421,0) rectangle (axis cs:82.8947368421053,26);
\draw[fill=color0,draw opacity=0] (axis cs:87.6315789473684,0) rectangle (axis cs:92.3684210526316,287);
\draw[fill=color0,draw opacity=0] (axis cs:97.1052631578947,0) rectangle (axis cs:101.842105263158,58);
\draw[fill=color0,draw opacity=0] (axis cs:106.578947368421,0) rectangle (axis cs:111.315789473684,24);
\draw[fill=color0,draw opacity=0] (axis cs:116.052631578947,0) rectangle (axis cs:120.789473684211,9);
\draw[fill=color0,draw opacity=0] (axis cs:125.526315789474,0) rectangle (axis cs:130.263157894737,15);
\draw[fill=color0,draw opacity=0] (axis cs:135,0) rectangle (axis cs:139.736842105263,21);
\draw[fill=color0,draw opacity=0] (axis cs:144.473684210526,0) rectangle (axis cs:149.210526315789,43);
\draw[fill=color0,draw opacity=0] (axis cs:153.947368421053,0) rectangle (axis cs:158.684210526316,63);
\draw[fill=color0,draw opacity=0] (axis cs:163.421052631579,0) rectangle (axis cs:168.157894736842,96);
\draw[fill=color0,draw opacity=0] (axis cs:172.894736842105,0) rectangle (axis cs:177.631578947368,680);
\end{axis}

\end{tikzpicture}
\begin{tikzpicture}

\definecolor{color0}{rgb}{0.12156862745098,0.466666666666667,0.705882352941177}
\scriptsize
\begin{axis}[
xlabel={Translation err [mm]},
ylabel={views},
xmin=-3.55263157894737, xmax=103.552631578947,
ymin=0, ymax=199.5,
width=\columnwidth,
height=\columnwidth,
xtick={0,25,50,75,100},
tick align=outside,
tick pos=left,
x grid style={lightgray!92.02614379084967!black},
y grid style={lightgray!92.02614379084967!black}
]
\draw[fill=color0,draw opacity=0] (axis cs:1.31578947368421,0) rectangle (axis cs:3.94736842105263,108);
\draw[fill=color0,draw opacity=0] (axis cs:6.57894736842105,0) rectangle (axis cs:9.21052631578947,181);
\draw[fill=color0,draw opacity=0] (axis cs:11.8421052631579,0) rectangle (axis cs:14.4736842105263,190);
\draw[fill=color0,draw opacity=0] (axis cs:17.1052631578947,0) rectangle (axis cs:19.7368421052632,140);
\draw[fill=color0,draw opacity=0] (axis cs:22.3684210526316,0) rectangle (axis cs:25,118);
\draw[fill=color0,draw opacity=0] (axis cs:27.6315789473684,0) rectangle (axis cs:30.2631578947368,155);
\draw[fill=color0,draw opacity=0] (axis cs:32.8947368421053,0) rectangle (axis cs:35.5263157894737,83);
\draw[fill=color0,draw opacity=0] (axis cs:38.1578947368421,0) rectangle (axis cs:40.7894736842105,97);
\draw[fill=color0,draw opacity=0] (axis cs:43.421052631579,0) rectangle (axis cs:46.0526315789474,52);
\draw[fill=color0,draw opacity=0] (axis cs:48.6842105263158,0) rectangle (axis cs:51.3157894736842,46);
\draw[fill=color0,draw opacity=0] (axis cs:53.9473684210526,0) rectangle (axis cs:56.5789473684211,16);
\draw[fill=color0,draw opacity=0] (axis cs:59.2105263157895,0) rectangle (axis cs:61.8421052631579,28);
\draw[fill=color0,draw opacity=0] (axis cs:64.4736842105263,0) rectangle (axis cs:67.1052631578948,17);
\draw[fill=color0,draw opacity=0] (axis cs:69.7368421052632,0) rectangle (axis cs:72.3684210526316,14);
\draw[fill=color0,draw opacity=0] (axis cs:75,0) rectangle (axis cs:77.6315789473684,7);
\draw[fill=color0,draw opacity=0] (axis cs:80.2631578947368,0) rectangle (axis cs:82.8947368421053,8);
\draw[fill=color0,draw opacity=0] (axis cs:85.5263157894737,0) rectangle (axis cs:88.1578947368421,6);
\draw[fill=color0,draw opacity=0] (axis cs:90.7894736842105,0) rectangle (axis cs:93.421052631579,9);
\draw[fill=color0,draw opacity=0] (axis cs:96.0526315789474,0) rectangle (axis cs:98.6842105263158,5);
\end{axis}

\end{tikzpicture}
		\end{minipage}%
		\begin{minipage}[t]{0.5\columnwidth}
\begin{tikzpicture}

\definecolor{color0}{rgb}{0.12156862745098,0.466666666666667,0.705882352941177}
\scriptsize
\begin{axis}[
xlabel={Rotation err [deg]},
xmin=0, xmax=180,
ymin=0, ymax=710.85,
width=\columnwidth,
height=\columnwidth,
tick align=outside,
tick pos=left,
x grid style={lightgray!92.02614379084967!black},
y grid style={lightgray!92.02614379084967!black}
]
\draw[fill=color0,draw opacity=0] (axis cs:2.36842105263158,0) rectangle (axis cs:7.10526315789474,334);
\draw[fill=color0,draw opacity=0] (axis cs:11.8421052631579,0) rectangle (axis cs:16.5789473684211,67);
\draw[fill=color0,draw opacity=0] (axis cs:21.3157894736842,0) rectangle (axis cs:26.0526315789474,35);
\draw[fill=color0,draw opacity=0] (axis cs:30.7894736842105,0) rectangle (axis cs:35.5263157894737,6);
\draw[fill=color0,draw opacity=0] (axis cs:40.2631578947368,0) rectangle (axis cs:45,7);
\draw[fill=color0,draw opacity=0] (axis cs:49.7368421052632,0) rectangle (axis cs:54.4736842105263,13);
\draw[fill=color0,draw opacity=0] (axis cs:59.2105263157895,0) rectangle (axis cs:63.9473684210526,26);
\draw[fill=color0,draw opacity=0] (axis cs:68.6842105263158,0) rectangle (axis cs:73.421052631579,47);
\draw[fill=color0,draw opacity=0] (axis cs:78.1578947368421,0) rectangle (axis cs:82.8947368421053,52);
\draw[fill=color0,draw opacity=0] (axis cs:87.6315789473684,0) rectangle (axis cs:92.3684210526316,245);
\draw[fill=color0,draw opacity=0] (axis cs:97.1052631578947,0) rectangle (axis cs:101.842105263158,67);
\draw[fill=color0,draw opacity=0] (axis cs:106.578947368421,0) rectangle (axis cs:111.315789473684,36);
\draw[fill=color0,draw opacity=0] (axis cs:116.052631578947,0) rectangle (axis cs:120.789473684211,8);
\draw[fill=color0,draw opacity=0] (axis cs:125.526315789474,0) rectangle (axis cs:130.263157894737,16);
\draw[fill=color0,draw opacity=0] (axis cs:135,0) rectangle (axis cs:139.736842105263,25);
\draw[fill=color0,draw opacity=0] (axis cs:144.473684210526,0) rectangle (axis cs:149.210526315789,42);
\draw[fill=color0,draw opacity=0] (axis cs:153.947368421053,0) rectangle (axis cs:158.684210526316,59);
\draw[fill=color0,draw opacity=0] (axis cs:163.421052631579,0) rectangle (axis cs:168.157894736842,101);
\draw[fill=color0,draw opacity=0] (axis cs:172.894736842105,0) rectangle (axis cs:177.631578947368,677);
\end{axis}

\end{tikzpicture}
\begin{tikzpicture}

\definecolor{color0}{rgb}{0.12156862745098,0.466666666666667,0.705882352941177}
\scriptsize
\begin{axis}[
xlabel={Translation err [mm]},
xmin=-3.55263157894737, xmax=103.552631578947,
ymin=0, ymax=978.6,
width=\columnwidth,
height=\columnwidth,
xtick={0,25,50,75,100},
tick align=outside,
tick pos=left,
x grid style={lightgray!92.02614379084967!black},
y grid style={lightgray!92.02614379084967!black}
]
\draw[fill=color0,draw opacity=0] (axis cs:1.31578947368421,0) rectangle (axis cs:3.94736842105263,932);
\draw[fill=color0,draw opacity=0] (axis cs:6.57894736842105,0) rectangle (axis cs:9.21052631578947,242);
\draw[fill=color0,draw opacity=0] (axis cs:11.8421052631579,0) rectangle (axis cs:14.4736842105263,51);
\draw[fill=color0,draw opacity=0] (axis cs:17.1052631578947,0) rectangle (axis cs:19.7368421052632,15);
\draw[fill=color0,draw opacity=0] (axis cs:22.3684210526316,0) rectangle (axis cs:25,5);
\draw[fill=color0,draw opacity=0] (axis cs:27.6315789473684,0) rectangle (axis cs:30.2631578947368,3);
\draw[fill=color0,draw opacity=0] (axis cs:32.8947368421053,0) rectangle (axis cs:35.5263157894737,7);
\draw[fill=color0,draw opacity=0] (axis cs:38.1578947368421,0) rectangle (axis cs:40.7894736842105,5);
\draw[fill=color0,draw opacity=0] (axis cs:43.421052631579,0) rectangle (axis cs:46.0526315789474,1);
\draw[fill=color0,draw opacity=0] (axis cs:48.6842105263158,0) rectangle (axis cs:51.3157894736842,8);
\draw[fill=color0,draw opacity=0] (axis cs:53.9473684210526,0) rectangle (axis cs:56.5789473684211,0);
\draw[fill=color0,draw opacity=0] (axis cs:59.2105263157895,0) rectangle (axis cs:61.8421052631579,7);
\draw[fill=color0,draw opacity=0] (axis cs:64.4736842105263,0) rectangle (axis cs:67.1052631578948,3);
\draw[fill=color0,draw opacity=0] (axis cs:69.7368421052632,0) rectangle (axis cs:72.3684210526316,5);
\draw[fill=color0,draw opacity=0] (axis cs:75,0) rectangle (axis cs:77.6315789473684,2);
\draw[fill=color0,draw opacity=0] (axis cs:80.2631578947368,0) rectangle (axis cs:82.8947368421053,3);
\draw[fill=color0,draw opacity=0] (axis cs:85.5263157894737,0) rectangle (axis cs:88.1578947368421,3);
\draw[fill=color0,draw opacity=0] (axis cs:90.7894736842105,0) rectangle (axis cs:93.421052631579,3);
\draw[fill=color0,draw opacity=0] (axis cs:96.0526315789474,0) rectangle (axis cs:98.6842105263158,3);
\end{axis}

\end{tikzpicture}
	\end{minipage}}
	\caption{Rotation and translation error histograms on all T-LESS test scenes with our RGB-based (left columns) and ICP-refined (right columns) 6D Object Detection }
	\label{fig:histo}
\end{figure*}
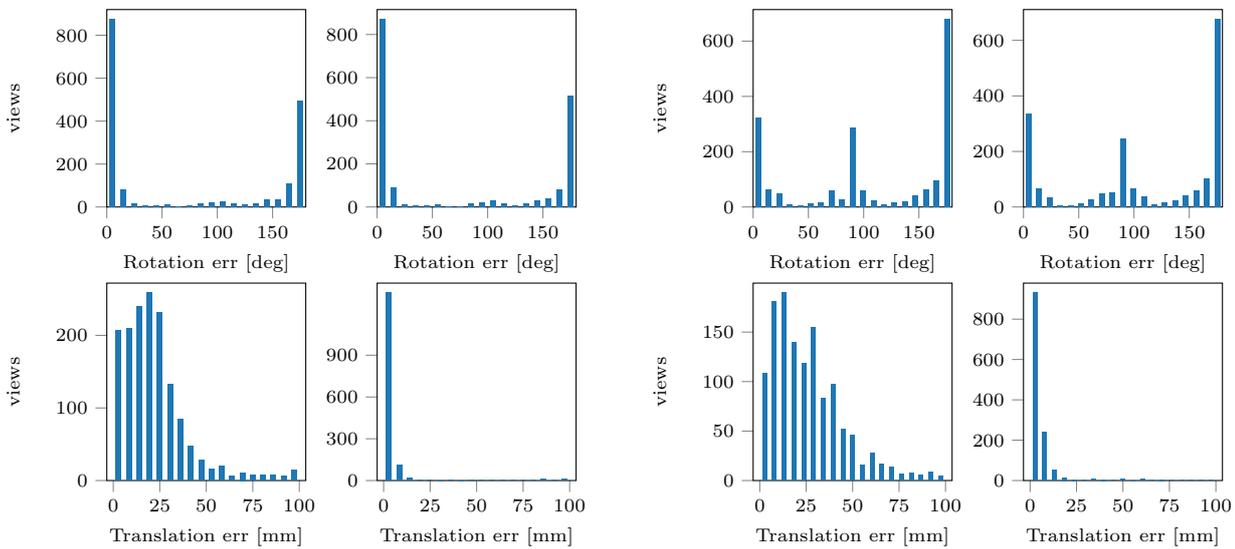
\begin{figure*}[t]
	\centering
	\subfloat[]{{\includegraphics[width=0.31\linewidth]{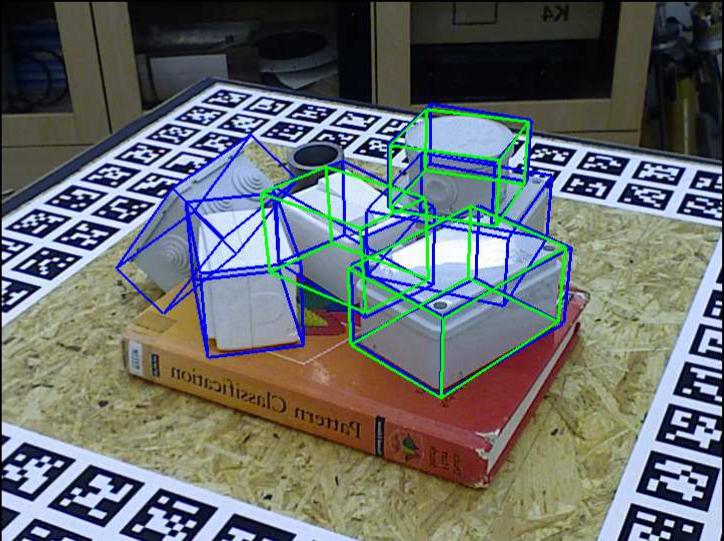} }}%
	\subfloat[]{{\includegraphics[width=0.31\linewidth]{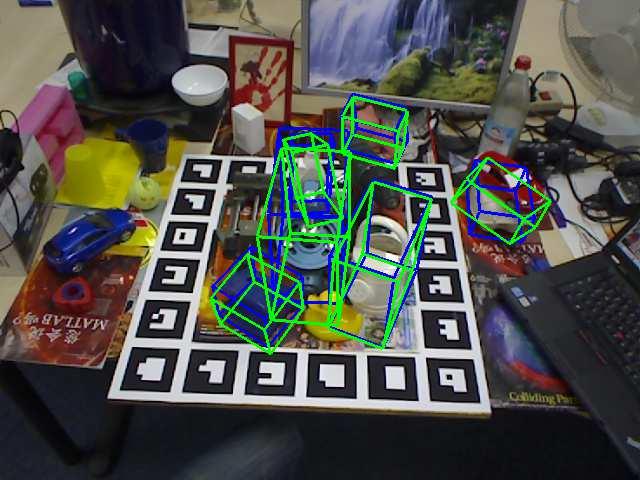} }}%
	\subfloat[]{{\includegraphics[width=0.31\linewidth]{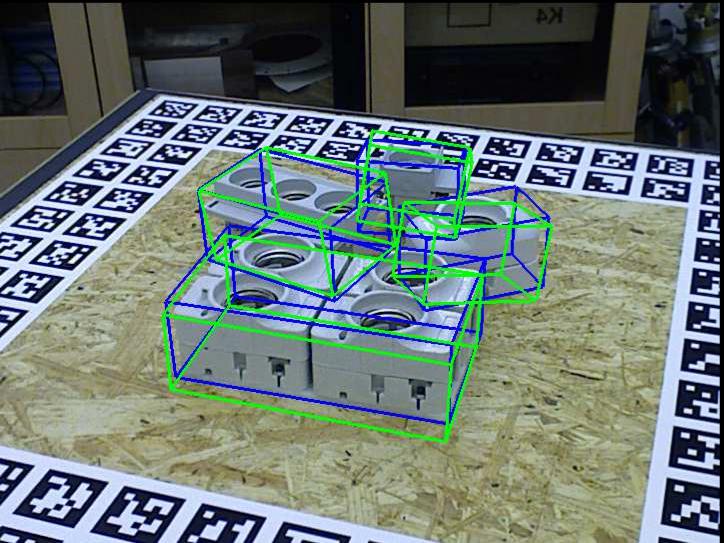} }}
	\caption{Failure cases; Blue: True poses; Green: Predictions; (a) Failed detections due to occlusions and object ambiguity, (b) failed AAE predictions of Glue (middle) and Eggbox (right) due to strong occlusion, (c) inaccurate predictions due to occlusion}
	\label{fig:failures}%
\end{figure*}
	\subsection{Discussion of 6D Object Detection results}
	Our RGB-only 6D Object Detection pipeline consists of 2D detection, 3D orientation estimation, projective distance estimation and perspective error correction. Although the results are visually appealing, to reach the performance of state-of-the-art depth-based methods we also need to refine our estimates using a depth-based \gls{ICP}. 
	Table \ref{tab:tless} presents our 6D detection evaluation on all scenes of the T-LESS dataset, which contains a high amount of pose ambiguities. Our pipeline outperforms all 15 reported T-LESS results on the 2018 BOP benchmark from \cite{hodan2018bop} in a fraction of the runtime. Table \ref{tab:tless} shows an extract of competing methods. Our RGB-only results can compete with the RGB-D learning-based approaches of \cite{brachmann2016uncertainty} and \cite{kehl2016deep}. Previous state-of-the-art approaches from \cite{vidal20186d,drost2010model} perform a time consuming refinement search through multiple pose hypotheses while we only perform the ICP on a single pose hypothesis. That being said, the codebook is well suited to generate multiple hypotheses using $k>1$ nearest neighbors. The right part of Table \ref{tab:tless} shows results with ground truth bounding boxes yielding an upper bound on the pose estimation performance.
	
	The results in Table \ref{tab:tless} show that our domain randomization strategy allows to generalize from 3D reconstructions as well as untextured CAD models as long as the considered objects are not significantly textured. Instead of a performance drop we report an increased \gls{VSD}$<0.3$ recall due to the more accurate geometry of the model which results in correct bounding box diagonals and thus a better projective distance estimation in the RGB-domain.
	
	In Table \ref{tab:linemod} we also compare our pipeline against state-of-the-art methods on the LineMOD dataset. Here, our synthetically trained pipeline does not reach the performance of approaches that use real pose annotated training data.

	There are multiple issues: (1) As described in Sec \ref{sec:cond} the real training and test set are strongly correlated and approaches using the real training set can over-fit to it; (2) the models provided in LineMOD are quite bad which affects both, the detection and pose estimation performance of synthetically trained approaches; (3) the advantage of not suffering from pose-ambiguities does not matter much in LineMOD where most object views are pose-ambiguity free; (4) We train and test poses from the whole SO(3) as opposed to only a limited range in which the test poses lie.
	SSD6D also trains only on synthetic views of the 3D models and we outperform their approach by a big margin in the RGB-only domain before \gls{ICP} refinement. 
	
	\subsection{Failure Cases}
	Figure \ref{fig:failures} shows qualitative failure cases, mostly stemming from missed detections and strong occlusions. A weak point is the dependence on the bounding box size at test time to predict the object distance. Specifically, under sever occlusions the predicted bounding box tends to shrink such that it does not encompass the occluded parts of the detected object even if it is trained to do so.
	If the usage of depth data is clear in advance other methods for directly using depth-based methods for distance estimation might be better suited. Furthermore, on strongly textured objects, the AAEs should not be trained without rendering the texture since otherwise the texture might not be distinguishable from shape at test time. The sim2real transfer on strongly reflective objects like satellites can be challenging and might require physically-based renderings.
	Some objects, like long, thin pens can fail because their tight object crops at training and test time appear very near from some views and very far from other views, thus hindering the learning of proper pose representations. As the object size is unknown during test time, we cannot simply crop a constantly sized area.

\begin{figure}[t]
	\centering
	\captionsetup{width=0.99\columnwidth}
	\includegraphics[width=0.9\columnwidth]{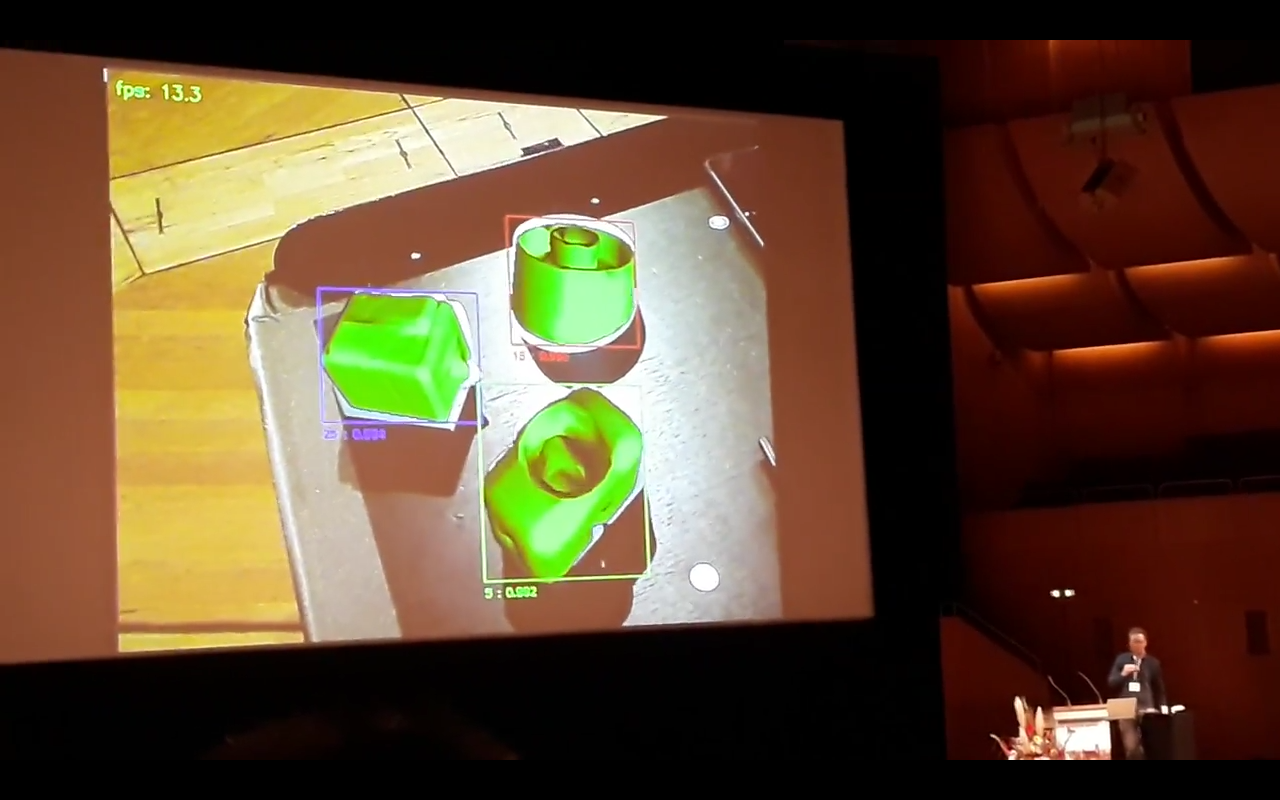}
	\caption{MobileNetSSD and \glspl{AAE} on T-LESS objects, demonstrated live at ECCV 2018 on a Jetson TX2}
	\label{fig:ECCV-live-demo}
\end{figure}

	\subsection{Rotation and Translation Histograms}
	
	To investigate the effect of \gls{ICP} and to obtain an intuition about the pose errors, we plot the rotation and translation error histograms of two T-LESS objects (Fig. \ref{fig:histo}). We can see the view-dependent symmetry axes of both objects in the rotation errors histograms. We also observe that the translation error is strongly improved through the depth-based ICP while the rotation estimates from the \gls{AAE} are hardly refined. Especially when objects are partly occluded, the bounding boxes can become inaccurate and the projective distance estimation (Sec. \ref{sec:proj}) fails to produce very accurate distance predictions. Still, our global and fast 6D Object Detection provides sufficient accuracy for an iterative local refinement method to reliably converge.
	
	\subsection{Demonstration on Embedded Hardware}
	
	The presented \glspl{AAE} were also ported onto a Nvidia Jetson TX2 board, together with a small footprint MobileNet from \cite{howard2017mobilenets} for the bounding box detection. A webcam was connected, and this setup was demonstrated live at ECCV 2018, both in the demo session and during the oral presentation. For this demo we acquired several of the T-LESS objects. As can be seen in Figure~\ref{fig:ECCV-live-demo}, lighting conditions were dramatically different than in the test sequences from the T-LESS dataset which validates the robustness and applicability of our approach outside lab conditions. No ICP was used, so the errors in depth resulting from the scaling errors of the MobileNet, were not corrected. However, since small errors along the depth direction are less perceptible for humans, our approach could be interesting for augmented reality applications. The detection, pose estimation and visualization of the three test objects ran at over 13Hz.
	
	\section{Conclusion}
	We have proposed a new self-supervised training strategy for Autoencoder architectures that enables robust 3D object orientation estimation on various RGB sensors while training only on synthetic views of a 3D model. By demanding the Autoencoder to revert geometric and color input augmentations, we learn representations that (1) specifically encode 3D object orientations, (2) are invariant to a significant domain gap between synthetic and real RGB images, (3) inherently regard pose ambiguities from symmetric object views. Around this approach, we created a real-time (42 fps), RGB-based pipeline for 6D object detection which is especially suitable when pose-annotated RGB sensor data is not available.

	\section*{Acknowledgement}
	We would like to thank Dr. Ingo Kossyk, Dimitri Henkel and Max Denninger for helpful discussions. We also thank the reviewers for their useful comments. This work has been partly funded by Robert Bosch GmbH, Corporate Research.

\bibliographystyle{spbasic}      
\bibliography{refs}   

\end{document}